\documentclass[11pt, a4paper, logo, copyright]{googledeepmind}

\usepackage[authoryear, sort&compress, round]{natbib}
\title{NExT: Teaching Large Language Models to Reason about Code Execution}

\correspondingauthor{Ansong Ni  $\langle$ansong.ni@yale.edu$\rangle$, Pengcheng Yin $\langle$pcyin@google.com$\rangle$}

\author[1 2]{Ansong Ni}
\author[1]{Miltiadis Allamanis}
\author[2]{Arman Cohan}
\author[1 3]{Yinlin Deng}
\author[1]{Kensen Shi}
\author[1]{Charles Sutton}
\author[1]{Pengcheng Yin}

\affil[1]{Google DeepMind}
\affil[2]{Yale University}
\affil[3]{University of Illinois at Urbana-Champaign}

\usepackage{microtype}
\usepackage{graphicx}
\usepackage{booktabs} %

\usepackage{mathtools}
\usepackage{amsthm}
\usepackage{amsfonts, amsmath, amssymb}
\usepackage{longtable}

\usepackage[nameinlink,capitalize]{cleveref}

\usepackage{listings}
\usepackage{algorithmic}
\usepackage{algorithm}
\usepackage{colortbl}
\usepackage{xcolor}
\usepackage{float}
\usepackage{amsmath}
\usepackage{amssymb}
\usepackage{graphicx}
\usepackage{booktabs}
\usepackage{bbm}
\usepackage{bm}
\usepackage{soul}
\usepackage{caption}
\usepackage{subcaption}
\usepackage{multirow}
\usepackage{arydshln}
\usepackage[normalem]{ulem}
\usepackage{xspace}
\usepackage{helvet}
\usepackage{tikz}
\usepackage[frozencache,cachedir=.]{minted}
\usepackage{soul}

\setminted{fontsize=\scriptsize}
\newenvironment{promptfence}{\captionsetup{type=listing}}{}

\useunder{\uline}{\ul}{}

\def\arcade/{\textsc{Arcade}}
\def\dsonek/{\textsc{DS-1000}}
\def\mbpp/{\textsc{Mbpp}}
\def\mbppr/{\textsc{Mbpp-R}}
\def\he/{\textsc{HumanEval}}
\def\hepackfull/{\textsc{Human\-Eval\-Fix-Plus}}
\def\hepack/{\textsc{He\-Fix+}}

\makeatletter
\Crefname{algorithm}{Algo.}{Algorithms}
\Crefname{table}{Tab.}{Tables}
\crefname{section}{\S\@gobble}{\S\S\@gobble}
\crefname{subsection}{\S\@gobble}{\S\S\@gobble}
\makeatother

\newcommand{\code}[1]{{\tt {\small #1}}}
\newcommand{\patk}{\textsc{pass}@$k$\xspace}
\newcommand{\patks}[1]{\textsc{pass}@$#1$\xspace}
\newcommand{\eg}{\textit{e.g., \xspace}}
\newcommand{\ie}{\textit{i.e., \xspace}}

\newcommand{\x}{\ensuremath{x}}
\newcommand{\hy}{\ensuremath{\hat{y}}}

\newcommand{\hr}{\ensuremath{\hat{r}}}
\newcommand{\buggy}{\ensuremath{\tilde{y}}}
\newcommand{\rat}{\ensuremath{r}}
\newcommand{\y}{\ensuremath{y}}

\newcommand{\bt}[1]{\textcolor{blue}{\textit{#1}}}

\newcommand{\lm}{P_{\theta}}

\newcommand{\lmi}[1]{P_{\theta^{#1}}}
\newcommand{\bfr}{\mathcal{B}}
\newcommand{\exe}{\mathcal{E}}
\newcommand{\tc}{T}
\newcommand{\tr}{\epsilon}
\newcommand{\fullOurs}{Naturalized Execution Tuning}
\newcommand{\ours}{NE\textsc{x}T\xspace}
\newcommand{\palm}{PaLM~2\xspace}
\newcommand{\ourpalm}{PaLM~2-L$+$\textsc{NExT}\xspace}

\newcommand{\hide}[1]{}

\definecolor{gyellow}{HTML}{FCF2D1}
\definecolor{gpurple}{HTML}{603CAF}
\newcommand{\ghl}[1]{{\sethlcolor{gyellow}\color{gpurple}\hl{\textbf{\small #1}}}}

\renewcommand{\tt}[1]{\fontfamily{cmtt}\selectfont #1}

\newcommand\mybox[2][]{\tikz[overlay]\node[inner sep=1pt, anchor=text, rectangle, rounded corners=1mm,#1] {#2};\phantom{#2}}
\definecolor{fillcolor}{RGB}{216,217,252}
\newcommand\goodratspan[1]{\mybox[fill=cyan!15]{#1}}
\newcommand\badratspan[1]{\mybox[fill=red!15]{#1}}
\newcommand\tracespan[1]{\mybox[fill=yellow!20]{#1}}
\newcommand\trace[1]{{\tt \textcolor{magenta}{Trace #1}}}

\newcommand{\vcenteredinclude}[1]{\begingroup
\setbox0=\hbox{\includegraphics[height=1.0em]{#1}}%
\parbox{\wd0}{\box0}\endgroup}

\newcommand{\xxmark}{\vcenteredinclude{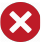}}
\newcommand{\ccmark}{\vcenteredinclude{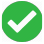}}
\newcommand{\ycmark}{\vcenteredinclude{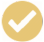}}

\theoremstyle{plain}

\theoremstyle{definition}

\theoremstyle{remark}

\newcommand{\alglinelabel}{%
  \addtocounter{ALC@line}{-1}%
  \refstepcounter{ALC@line}%
  \label%
}

\begin{abstract}
A fundamental skill among human developers is the ability to understand and reason about program execution.
As an example, a programmer can mentally simulate code execution in natural language to debug and repair code (\textit{aka}.~rubber duck debugging).
However, large language models (LLMs) of code are typically trained on the surface textual form of programs, thus may lack a semantic understanding of how programs execute at run-time.
To address this issue, we propose \ours, a method to teach LLMs to inspect the execution traces of programs (variable states of executed lines) and reason about their run-time behavior through chain-of-thought (CoT) rationales.
Specifically, \ours uses self-training to bootstrap a synthetic training set of execution-aware rationales that lead to correct task solutions (\eg fixed programs) without laborious manual annotation.
Experiments on program repair tasks based on \mbpp/ and \textsc{HumanEval} demonstrate that \ours improves the fix rate of a \palm model, by {$\mathbf{26.1\%}$} and {$\mathbf{14.3\%}$} absolute, respectively, with significantly improved rationale quality as verified by automated metrics and human raters.
Our model can also generalize to scenarios where program traces are absent at test-time. 
\end{abstract}

\begin{document}

\maketitle

\section{Introduction}

Recent years have witnessed the burgeoning of large language models (LLMs) trained on code~\citep{austin2021program,chen2021evaluating,anil2023palm,touvron2023llama,li2023starcoder,roziere2023code}.
While those LLMs achieve impressive performance in assisting developers with writing~\citep{chen2021evaluating}, editing~\citep{Fakhoury2023TowardsGF}, explaining~\citep{Hu2018DeepCC}, and reviewing~\citep{Li2022AutomatingCR} code, they still struggle on more complex software engineering tasks that require reasoning about the runtime execution behavior of programs~\citep{Ma2023ChatGPTUC}.
On the other hand, it is not always sufficient
for the model to suggest good code solutions, but it is 
often necessary to provide an explanation to developers
to document what the change does and why it is needed.
These explanations can help developers better understand the code solutions from models and make more informative decisions.  \citep{Cito2022counterfactual,Ross2023ThePA,Kang2023ExplainableAD}.

For example, \emph{program repair}~\citep{Chen2018SequenceRSL,Li2020DLFixCC,goues2019automated} is the task of fixing bugs in a program.
Human developers usually learn to debug and fix code by interacting with code interpreters or debuggers to inspect the variable states of executed lines~\citep{siegmund14studying}.
Such practice helps them acquire a \textit{mental model} of program execution~\citep{Heinonen2022SynthesizingRO}, so that they could mentally simulate code execution in a more abstract manner using natural language as in  rubber duck debugging~\citep{Hunt1999ThePP}.
Therefore, a program repair model would be more helpful to developers if the model could carry out similar reasoning about program execution in order to explain bugs to programmers.

With this inspiration, our goal is to improve the ability of LLMs to reason about program execution when solving coding tasks. In this paper we propose \fullOurs\ (\ours), which aims to teach LLMs to reason with code execution by inspecting program execution traces and reasoning about the code's runtime behavior in natural language (NL).
At a general level, for a coding task,
the main idea is to train a model to generate intermediate NL rationales, as in chain-of-thought reasoning~\citep{chainofthought}, but to provide
the model with a trace of the execution of the program
in question,
so the rationale can be more accurate and grounded on program semantics.
Teaching LLMs to reason about program execution in NL would not only offer better interpretability, it could also increase the diversity of  solutions predicted by the model~\citep{yin2023arcade}.

\begin{figure*}
    \centering
    \includegraphics[width=\textwidth]{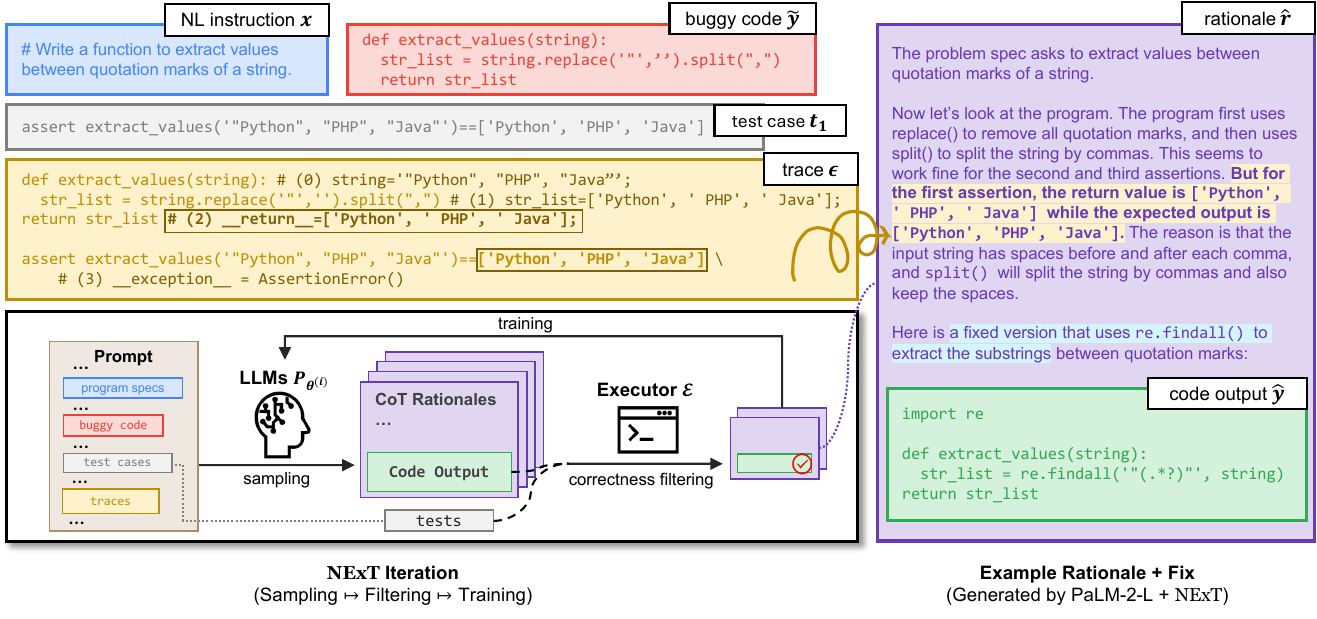}
    \caption{\ours finetunes an LLM to \textit{naturalize} execution traces into the chain-of-thought rationales for solving coding tasks. It performs \textit{iterative self-training} from weak supervision, by learning from samples that lead to correct task solutions.
    }
    \label{fig:next}
\vspace{-3mm}
\end{figure*}

\cref{fig:next} illustrates our proposed approach when applied to program repair.
Given an NL task instruction ($\x$ in \cref{fig:next}) and a buggy program ($\buggy$), 
as well as the execution traces of the program ($\tr$), an LLM solves the task (\eg predict the fixed code $\hy$) using chain-of-thought (CoT) reasoning to generate a \emph{natural language rationale} ($\hat{\rat}$) leveraging the execution information\footnote{While there are a variety types of execution information that we may provide to an LLM (\eg variable read/write, runtime environments), in this work we limit the execution information to program states and variable values from the execution trace, which is common information that (human) developers also use.}.
Intuitively, program traces encode useful debugging information such as line-by-line variable states (\eg the value of \code{str\_list} in $\tr$, \cref{fig:next}) or any exceptions thrown, which could be useful for LLMs to identify and fix bugs by reasoning over the expected and the actual execution results (\eg  \ghl{\textit{``highlighted text''}} in $\hat{\rat}$).
To help LLMs understand execution traces, \ours represent traces as compact inline code comments  (\eg~\tracespan{\code{\# (1) str\_list=$\ldots$}} in $\tr$, more in \cref{sec:pre_study}), without interrupting the original program structure.

While execution traces capture informative runtime behavior, we find it challenging for LLMs to effectively leverage them out-of-box through CoT prompting (\cref{sec:pre_study}).
Therefore we opt to finetune LLMs on high-quality CoT rationales that reason about program execution (\cref{sec:methodology}).
\ours uses weakly-supervised self-training~\citep{zelikman2022star} to bootstrap a synthetic training set by sampling rationales that lead to correct task solutions (\eg fixed code $\hy$ in \cref{fig:next}) verified by unit tests \citep[][]{Ye2022SelfAPRSP}.
Using unit tests as weak supervision, \ours learns to discover task-specific, execution-aware NL rationales without relying on laborious manual annotation of rationales~\citep{Chung2022ScalingIL,Longpre2023TheFC,lightman2023let} or distilling such data from stronger teacher models~\citep{Gunasekar2023TextbooksAA,Mukherjee2023OrcaPL,Mitra2023Orca2T,Fu2023SpecializingSL}.
\ours executes this self-training loop for multiple iterations~\citep{Anthony2017ThinkingFA,Dasigi2019IterativeSF}, solving more challenging tasks with improved success rate and rationale quality (\cref{sec:exp}).

We evaluate \ours with the \palm-L model~\citep{anil2023palm} on two Python program repair tasks.
Experiments (\cref{sec:exp}) show that \ours significantly improves \palm's ability to reason about program execution in natural language, improving the program fix rate on \mbppr/ by $26.1\%$ and \hepackfull/ by $14.3\%$ absolute, respectively.
When compared against a strong self-training program repair approach without predicting NL rationales~\citep{Ye2022SelfAPRSP}, our model achieves comparable accuracy with significantly improved sample diversity.
Interestingly, while our model learns to reason with pre-existing execution information in input program traces, it also generalizes to the out-of-distribution scenario where execution traces are not available at test-time.
Finally, to measure the quality of model-generated rationales, we propose a \textit{proxy-based} evaluation approach, which approximates rationale quality using the performance of smaller LLMs when prompted to solve the original task following those rationales from our models.
Through both proxy-based evaluation and human annotation, we demonstrate that \ours produces helpful NL rationales which explain the causes of bugs while suggesting potential fixes.
The generated rationales are of significantly higher quality compared to those from the base \palm-L model.

\definecolor{codegreen}{rgb}{0,0.6,0}
\definecolor{codegray}{rgb}{0.5,0.5,0.5}
\definecolor{codepurple}{rgb}{0.58,0,0.82}

\lstdefinestyle{tracing}{
    backgroundcolor=\color{white},   
    commentstyle=\color{magenta},
    keywordstyle=\color{blue},
    numberstyle=\bfseries\color{codegray}\fontsize{7.2pt}{8.5pt}\selectfont,
    stringstyle=\color{codepurple},
    basicstyle=\bfseries\ttfamily\fontsize{7.2pt}{8.5pt}\selectfont,
    breakatwhitespace=false,         
    breaklines=true,                 
    captionpos=b,                    
    keepspaces=true,                 
    numbers=left,                    
    numbersep=5pt,                  
    showspaces=false,                
    showstringspaces=false,
    showtabs=false,                  
    tabsize=2
}

\begin{figure*}
\centering
    \begin{tabular}{c}
    \toprule
        \begin{lstlisting}[style=tracing,language=Python]
def separate_odd_and_even(lst):  # (0) lst=[1, 2, 3, 4, 5, 6, 7, 8, 9, 10];
    odd_list = []  # (1) odd_list=[];
    even_list = []  # (2) even_list=[];
    for n in lst:  # (3) n=1; (5) n=2; (7) n=3; ...; (21) n=10;
       if n %
          even_list.append(n)  # (4) even_list=[1]; (8) even_list=[1, 3]; ...; (20) even_list=[1, 3, 5, 7, 9];
       else:
         odd_list.append(n)  # (6) odd_list=[2]; (10) odd_list=[2, 4]; ...; (22) odd_list=[2, 4, 6, 8, 10];
    return odd_list, even_list  # (23) __return__=([2, 4, 6, 8, 10], [1, 3, 5, 7, 9])
    
separate_odd_and_even([1,2,3,4,5,6,7,8,9,10]) == [1,3,5,7,9], [2,4,6,8,10]\end{lstlisting}
    \\\bottomrule
    \end{tabular}
    \caption{\ours represents execution trace as \texttt{{\color{magenta} \textbf{\footnotesize inline comments}}}. More details in \cref{sec:problem_statement} and \cref{sec:trace-repr-details}.
    }
    \label{fig:trace-example}
\end{figure*}

\section{Task: Program Repair with Traces}
\label{sec:problem_statement}

Here we introduce our task of program repair with execution traces using chain-of-thought reasoning.

\textbf{Program Repair with Execution Traces.}\quad
As in \cref{fig:next}, given an instruction $\x$ and a buggy code solution $\buggy$, automated program repair \citep{goues2019automated} aims to generate a fixed program $\hy$ such that $\hy$ passes all test cases $t \in \tc$ in an executor $\exe$, \ie $\exe (\hy, \tc) = 1$ while  $\exe (\buggy, \tc) = 0$.
In this paper we focus on the task of program repair using execution traces~\citep{Bouzenia2023TraceFixerET}.
Specifically, a \textbf{program trace} $\tr$ is a sequence of intermediate variable states after executing each statement in $\buggy$ against a test case $t$.
Intuitively, traces record the computation of a program, and can provide useful debugging information (\eg~exceptions) to repair $\buggy$.

To use LLMs to repair programs with traces, we concatenate the task instruction, the buggy code, the test cases, and their execution traces as a prompt (\cref{fig:next}). To help LLMs understand program traces, we design a prompt-friendly trace representation by formatting $\tr$ as compact inline code comments (\ie $\tr$ in \cref{fig:next}), as discussed later.

\textbf{CoT Reasoning with Execution.}\quad
We focus on using chain-of-thought reasoning \citep{wei2022chain} to solve program repair problems by reasoning with execution, where an LLM is prompted to generate an NL rationale $\hr$ together with a fixed program $\hy$ as in \cref{fig:next}.
Specifically, we consider rationales that contain reasoning steps to identify and explain bugs in the original code (\eg the second paragraph in $\hat{\rat}$, \cref{fig:next}), as well as suggestions to fix the buggy code (\eg \goodratspan{\textit{``a fixed version that uses}} \goodratspan{{\tt {\small re.findall()}}\textit{''}} in $\hat{\rat}$).
Since rationales are generated using traces, they often include useful reasoning about program execution that helps localize the bug, such as identifying a counterfactual between the expected and the actual variable values of a statement (\eg the \ghl{\textit{``highlighted text''}} in $\hat{\rat}$).
Such explanations can be helpful for developers to understand bugs in the original code and the model's fixed solutions~\citep{Kang2023ExplainableAD}.
We therefore aim to improve the quality of NL rationales along with the fix rate by teaching LLMs to reason with execution information.

\textbf{An LLM-friendly Trace Representation.}\quad
The raw execution traces collected at runtime contain complete variable states for each executed statement.\footnote{We use the {\tt sys.settrace()} hook in Python.}
Encoding all such information in prompts is not feasible given the context limit and computation overhead of LLMs.
To address this issue and make execution information more intelligible to LLMs, we propose an \textit{inline trace representation} format, which encodes variable states as inline comments of the traced program.
\cref{fig:trace-example} shows an example.
Specifically, each inline comment only encodes changed variables after executing that line. 
Because statements may be invoked multiple times in non-obvious orders (\eg in loops like lines 4 to 8 in \cref{fig:trace-example}), we index the variable states based on the execution order (\eg \code{(3) n=1;} and \code{(4) even\_list=[1]}), and one may reconstruct the original execution footprint by following those variable states in order. 
We further compress the trace information for loops by omitting the variable states in intermediate iterations (\eg ``\code{...}'' in lines 4, 6, and 8). 
Intuitively, by showing states as pseudo-comments within the original code without interrupting the program structure, our trace representation is significantly more compact than existing approaches that unroll executed lines of code and pair them with line-by-line variable states~\citep[\textit{c.f.},][]{nye2021show,Bouzenia2023TraceFixerET},\footnote{As a comparison, 95\% examples in our \mbppr/ benchmark can fit into a 2K context window using our inline representation, while only 60\% of them can fit into the same window using the Scratchpad trace format in \citet{nye2021show}. A more detailed comparison is shown in \cref{tab:trace_repr_length}.
}
while allowing an LLM to leverage its learned code representation to understand the additional execution effect of each statement.
Implementation details about handling complex control structures are discussed in~\cref{sec:trace-repr-details}.

\begin{table*}[t]
\centering
\footnotesize
\begin{tabular}{ll:cccccc:l}
\toprule
                                          &                                              &                                     &                                    &                                   & Mixtral                  & DeepSeek                 & StarCoder         & \multirow{2}{*}{\textbf{Avg.}}        \\
\multirow{-2}{*}{\textbf{Benchmarks}}         & \multirow{-2}{*}{\textbf{Prompting Methods}} & \multirow{-2}{*}{PaLM 2-L} & \multirow{-2}{*}{GPT-3.5} & \multirow{-2}{*}{GPT-4}  & 8x7B                     & Coder 33B                & 15.5B          &           \\\midrule
                           & Vanilla w/ trace         & 27.5                              & 41.8                              & 62.6                              & 16.1                              & 23.9                              & 13.3                              & 30.9                              \\
                           & \quad $+$ CoT            & {\color{red} {\ul 26.6}} & 46.4                              & 62.8                              & 21.1                              & {\color{red} {\ul 18.2}} & {\color{red} {\ul 12.6}} & 31.3$_{+0.4}$                            \\
\multirow{-3}{*}{\mbppr/}  & \quad $+$ CoT; $-$ trace & {\color{red} {\ul 19.0}} & 47.1                              & {\color{red} {\ul 51.3}} & {\color{red} {\ul 18.1}} & {\color{red} {\ul 12.9}} & {\color{red} {\ul 10.6}} & {\color{red} {\ul 26.5}$_{-4.8}$} \\ \midrule
                           & Vanilla w/ trace         & 59.1                              & 70.1                              & 88.4                              & 32.9                              & 57.3                              & 29.3                              & 56.2                              \\
                           & \quad $+$ CoT            & {\color{red} {\ul 48.8}} & 75.6                              & {\color{red} {\ul 84.8}} & 34.1                              & {\color{red} {\ul 30.5}} & {\color{red} {\ul 16.5}} & {\color{red} {\ul 48.4}$_{-7.8}$} \\
\multirow{-3}{*}{\hepack/} & \quad $+$ CoT; $-$ trace & {\color{red} {\ul 43.3}} & {\color{red} {\ul 72.0}} & {\color{red} {\ul 82.9}} & {\color{red} {\ul 25.6}} & {\color{red} {\ul 22.6}} & 18.3                              & {\color{red} {\ul 44.1}$_{-4.3}$} \\ \bottomrule
\end{tabular}
\caption{
Few(3)-shot prompting repair accuracy using greedy decoding.
Results worse than the previous row above them are {\color{red} {\ul underlined in red}}.
}
\label{tab:pre-study}
\end{table*}
\section{Preliminary Study: Can LLMs reason with program traces in natural language?}
\label{sec:pre_study}

Before introducing \ours, 
we first conduct a preliminary study to explore whether LLMs could reason with execution traces in natural language out-of-box without additional training.
Answering this question will motivate our finetuning approach to improve such reasoning skills. 
Specifically, we follow the trace representation in \cref{sec:problem_statement} and few-shot prompt an LLM to solve program repair tasks using CoT reasoning.

\textbf{Models.}\quad
We evaluate the following general-purpose models: \palm~\citep{anil2023palm}, GPT~\citep{openai2023gpt4}\footnote{We use {\tt gpt-3.5-turbo-1106} and {\tt gpt-4-1106-preview}.}, and Mixtral~\citep{jiang2024mixtral}.
We also test two code-specific LLMs: StarCoder~\citep{li2023starcoder} and DeepSeek Coder~\citep{guo2024deepseekcoder}.
\cref{tab:pre-study} reports the results on two Python program repair datasets (see \cref{sec:exp} for details). 

\textbf{LLMs struggle on CoT reasoning with traces.}\quad
We observed mixed results when comparing vanilla prompting with traces without CoT  (\textbf{Vanilla w/ trace} in \cref{tab:pre-study}) and CoT prompting with rationales ($\bm{+}$\textbf{CoT}).
Surprisingly, CoT prompting is even worse on \hepackfull/, with an average drop of $-7.8\%$ compared to vanilla prompting, especially for code-specific LLMs ($57.3\mapsto30.5$ for DeepSeek Coder and $29.3\mapsto16.5$ for StarCoder).
After inspecting sampled rationales predicted by \palm-L, we observe that the model is subject to strong hallucination issues, such as mentioning exceptions not reflected in the given traces.
Indeed, as we later show in \cref{sec:human_rating}, the overall correctness rate of explaining errors in input programs among these sampled rationales from \palm-L is only around $30\%$.
Moreover, CoT reasoning is even more challenging for those models when we remove execution traces from the inputs ($\bm{+}$\textbf{CoT;}$\bm{-}$\textbf{trace}), resulting in an average performance drop of $4.8\%$ on \mbppr/ and $4.3\%$ on \hepackfull/.
These results suggest that while our trace representation is useful for LLMs to understand and leverage execution information for program repair (since ``$-$trace'' leads to worse results), they could still fall short on CoT reasoning using natural language with those program traces.
This finding therefore motivates us to improve LLMs in reasoning with execution through finetuning, which we elaborate in \cref{sec:methodology}.

\section{\ours: Naturalized Execution Tuning}
\label{sec:methodology}
We present \ours, a self-training method to finetune LLMs to reason with program execution using synthetic rationales.

\textbf{Overview of \ours.}\quad
\cref{fig:next} illustrates \ours, with its algorithm detailed in \cref{alg:next-iter}.
\ours is based on existing self-trained reasoning approaches~\citep{zelikman2022star,Uesato2022SolvingMW}, which employ expert iteration to improve a base LLM using synthetic rationales sampled from the model.
Given a training set $\mathcal{D}$ of repair tasks with execution traces, \ours first samples candidate NL rationales and fixed code solutions from the LLM.
Those candidate solutions are filtered using unit test execution diagnostics, and those that pass all test cases are then used to update the model via finetuning.
This sample-filter-train loop is performed for multiple iterations, improving the model's rationales and repair success rate after each iteration.

\textbf{Sampling rationales and code solutions.}\quad
For each iteration $i$, we sample rationales $\hr$ and fixes $\hy$ in tandem from the current model $\lmi{(i)}$ (Line~\ref{alg:next:sampling}, \cref{alg:next-iter}).
We use few-shot prompting (\cref{sec:pre_study}) when $i=0$ and zero-shot prompting with trained models for later iterations.
In contrast to existing self-training methods that  leverage all training problems, \ours only samples candidate solutions from the subset of problems in $\mathcal{D}$ that are challenging for the base model $P_{\theta^{(0)}}$ to solve (Line \ref{alg:next:hardonly}).
Specifically, given a metric $\mathcal{M}(\cdot)$, we only use problems $d \in \mathcal{D}$ if $P_{\theta^{(0)}}$'s metric on $d$ is below a threshold $m$.
Refer to \cref{sec:exp} for more details about the $\mathcal{M}(\cdot)$ and $m$ of our program repair task.
Focusing on sampling solutions from those hard problems not only significantly reduces sampling cost, it also improves program repair accuracy, as it helps the model towards learning to solve more challenging problems. See \cref{sec:app_full_training_dev_results} for a more detailed analysis. 

\newlength{\textfloatsepsave}
\setlength{\textfloatsepsave}{\textfloatsep}
\setlength{\textfloatsep}{5pt}%

\begin{algorithm*}[!t]
\small
  \captionsetup{type=algorithm}\caption{Naturalized Execution Tuning (\ours)\label{alg:next-iter} 
  }
\begin{algorithmic}
    \STATE {\bfseries Input:} Training set $\mathcal{D} = \{(\x_j, \buggy_j, \tc_j, \tr_j)\}_{j=1}^{|\mathcal{D}|}$ (\cref{sec:problem_statement}); Development set $\mathcal{D}_{dev}$; Base LLM $\lmi{(0)}$; Number of iterations $I$; Executor $\exe$; Evaluation metric $\mathcal{M}$ and threshold $m$
\end{algorithmic}
\begin{algorithmic}[1]
    \STATE $\mathcal{D}_H \leftarrow \{d \mid d\in\mathcal{D}, \mathcal{M}(\lmi{(0)}, d) < m\}$\alglinelabel{alg:next:hardonly}   \;\bt{// Identify hard problems $\mathcal{D}_H$ with metric $\mathcal{M}(\cdot)<m$}
    \FOR{$i=0$ \textbf{to} $I$}
        \STATE $\bfr^{(i)}\leftarrow\{\}$ 
        \FOR{$(\x_j, \buggy_j, \tc_j, \tr_j)$ \textbf{in} $\mathcal{D}_H$}
            \STATE $S_j^{(i)} \sim \lmi{(i)}(\rat, \y \mid \x_j, \buggy_j, \tc_j, \tr_j)$ \alglinelabel{alg:next:sampling} \;\bt{// Sample rationales $\rat$ and fixes $\y$ using trace $\tr_j$.}
            \STATE $\bfr^{(i)} \leftarrow \bfr^{(i)}\cup\{(\hr, \hy) \mid (\hr, \hy) \in S_j^{(i)}, \exe(\hy, T_j) = 1 \}$ \alglinelabel{alg:next:filtering} \;\bt{// Filter with test cases $\tc_j$ and add to $\bfr^{(i)}$.}
        \ENDFOR
        \STATE $\theta^{(i+1)}\leftarrow \arg\max_{\theta}\mathbb{E}_{\bfr^{(i)}}[\lm(\hr, \hy \mid x, \buggy, T, \tr)]$ \alglinelabel{alg:next:tuning} \;\bt{// Finetune model $P_{\theta^{(0)}}$ with data in $\bfr^{(i)}$.}
    \ENDFOR
    \STATE $i^* \leftarrow \arg\max_{i}\sum_{d\sim\mathcal{D}_{dev}}\mathcal{M}(\lmi{(i)}, d) / |\mathcal{D}_{dev}|$ \;\bt{// Select the best checkpoint $i^*$}
\end{algorithmic}
\begin{algorithmic}
    \STATE {\bfseries Output:} model $\lmi{(i^*)}$
\end{algorithmic}
\end{algorithm*}

\setlength{\textfloatsep}{\textfloatsepsave}

\textbf{Filtering candidate solutions.}\quad
Given a candidate set of sampled NL rationales and their code fixes, \ours uses unit test execution results to identify plausible rationales that lead to correct fixes for learning (Line~\ref{alg:next:filtering}).
Using test execution diagnostics as a binary reward function is natural for program repair tasks since each repair problem in our dataset comes with unit tests to test the functional correctness of its proposed fixes~\citep{Ye2022SelfAPRSP}.
While we remark that this filtering criteria does not directly consider rationale quality, we empirically demonstrate in \cref{sec:exp} that the quality of rationales improves as learning continues.\footnote{The rationale and fix quality may plateau at a different iteration $i$.}

\textbf{Model training.}\quad
After collecting a set of training examples $\bfr^{(i)}$, we finetune the model to maximize the probability of generating the target rationales and code fixes given the task input (Line~\ref{alg:next:tuning}).
Following \citet{zelikman2022star}, 
we always finetune the model from its initial checkpoint $P_{\theta^{(0)}}$
to avoid over-fitting to instances sampled from early iterations that are potentially of lower-quality.

\textbf{Discussion.}\quad
\ours can be seen as an instantiation of the rationale bootstrapping method proposed in \citet{zelikman2022star} (\S~3.1), which synthesizes latent rationales with correct answers for math and logical reasoning tasks.
However, \ours focuses on program comprehension by reasoning with execution traces, which is critical for solving challenging coding tasks that require understanding execution information, such as program repair (\cref{sec:exp}).
Besides, \ours models both rationales and programs (code fixes) as latent variables.
Using unit test execution results as weak supervision, \ours is able to explore possible strategies to reason with execution and discover plausible rationales catered towards solving the specific downstream task.
As we show in \cref{sec:case-study}, rationales generated by \ours employ a variety of reasoning patterns to locate and explain bugs in our repair dataset.
Finally, while we apply \ours to program repair, our framework is general and can be extended to other programming tasks that require reasoning about execution, such as code generation with partial execution contexts \citep{yin2023arcade} or inferring program execution results \citep{nye2021show}, which we leave as important future work.

\section{Experiments}
\label{sec:exp}

\textbf{Models.}\quad
We evaluate \ours using \palm-L (Unicorn) as the base LLM~\citep{anil2023palm}. Its finetuning API is publicly accessible on Google Cloud Vertex AI platform.

\textbf{Datasets.}\quad
We use two Python program repair benchmarks, \textbf{\mbppr/} and \textbf{\hepackfull/} (\textbf{\hepack/} hereafter).
\mbppr/ is a new repair benchmark that we create from \mbpp/~\citep{austin2021program}, a popular function-level Python code generation dataset.
We create \mbppr/ by collecting LLM-generated incorrect code solutions to \mbpp/ problems, with a total of $10,047$ repair tasks for training and $1,468$ tasks (from a disjoint set of \mbpp/ problems) in the development for evaluation (\cref{sec:mbppr-details}).
In addition to \mbppr/, we also evaluate on \hepack/. \hepack/ is derived from \textsc{Human\-Eval\-Fix}~\citep{muennighoff2023octopack} which consists of 164 buggy programs for problems in the \he/ dataset~\citep{chen2021evaluating}. We further augment \textsc{Human\-Eval\-Fix} with the more rigorous test suites from EvalPlus~\citep{liu2023your} to obtain \hepack/.
While both original datasets \mbpp/ and \he/ feature function-level algorithmic code generation problems, problems from the two datasets may still differ in their topics, algorithms or data structures used.
Therefore, we use \hepack/ to measure generalization ability without further finetuning.

\begin{table*}[!t]
\centering
\scriptsize
\begin{tabular}{@{}lcccccccc@{}}
\toprule
                        & \multicolumn{4}{c}{\textbf{End-to-end Fix Rate}}                    & \multicolumn{4}{c}{\textbf{Proxy-based Evaluation {\scriptsize (\patks{k} on smaller LMs)}}}             \\\cmidrule(lr){2-5}\cmidrule(lr){6-9}
\textbf{Models}         & \patks{1}      & \patks{5}      & \patks{10}     & \patks{25}       & \patks{1}     & \patks{5}     & \patks{10}    & \patks{25}    \\\midrule
GPT-4; 3-shot           & 63.2           & 75.1           & 78.5           & 82.7             & 44.8          & 66.5          & 72.5          & 77.8          \\
GPT-3.5; 3-shot         & 42.9           & 65.0           & 70.7           & 76.7             & 26.6          & 48.8          & 57.0          & 66.4          \\
\palm-L; 3-shot        & 23.2           & 45.7           & 54.7           & 65.0             & 22.5          & 43.4          & 51.9          & 61.5          \\ \midrule
\ourpalm; 0-shot & 49.3$_{+26.1}$ & 68.1$_{+22.4}$ & 73.5$_{+18.8}$ & 79.4$_{+14.4}$   & 28.8$_{+6.3}$ & 49.9$_{+6.5}$ & 57.3$_{+5.4}$ & 65.5$_{+4.0}$ \\\bottomrule
\end{tabular}
\caption{Improvements by \ours on the \palm-L model (in subscripts) on \mbppr/. GPT-3.5/4 results are for reference.}
\label{tab:compare-gpt}
\end{table*}
\begin{figure*}[!t]
    \centering
    \includegraphics[width=\linewidth]{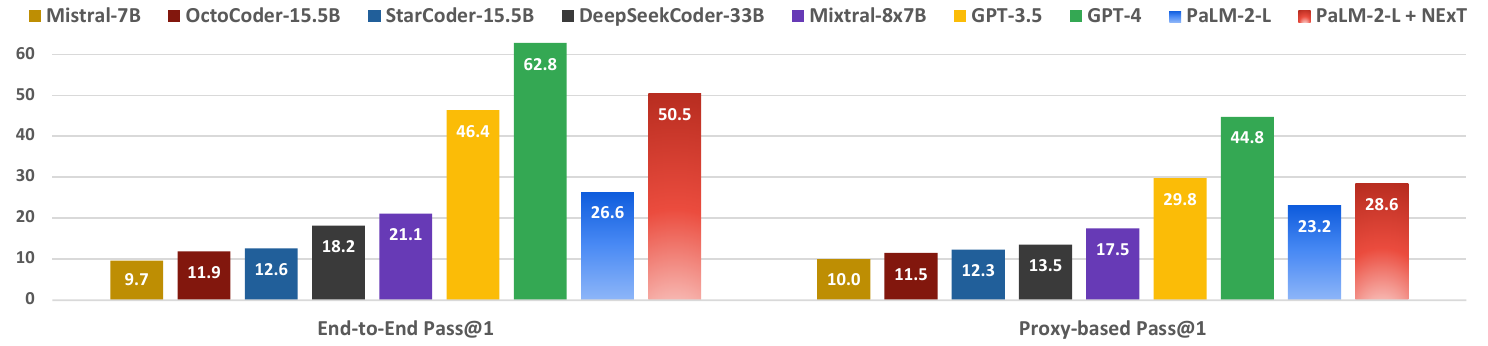}
    \caption{Greedy-decoding results on \mbppr/ on \ourpalm and existing LLMs.
    }
    \label{fig:external-comparison}
\end{figure*}

\textbf{Evaluating Code Fixes.}\quad
We use \patk~\citep{kulal2019spoc,chen2021evaluating}, defined as the fraction of solved repair tasks using $k$ samples ($k \leq 25$), to measure the end-to-end functional correctness of fixed programs with tests.

\textbf{Evaluating Rationale Quality.}\quad
Decoupling the quality of intermediate CoT rationales and downstream task performance (program repair \patk) is a non-trivial research question in LLM reasoning~\citep{prasad2023receval}, with most works on improving CoT reasoning still hill-climbing towards downstream task performance without evaluating intermediate rational quality (\eg \citet{lightman2023let}).
To disentangle the evaluation of rationale quality from end-to-end repair accuracy, we propose an extrinsic \textbf{proxy-based evaluation} metric for rationales.
Specifically, given a rationale $\rat$, we prompt a smaller LLM to solve the original repair task conditioning on $\rat$, and use the correctness of the predicted code fix (using greedy decoding) to approximate the quality of $\rat$.
Intuitively, smaller LLMs would rely more on information from the rationale and could be more sensitive to its errors.
Therefore, their performance could be a better indicator of rationale quality.
We report averaged scores on two \palm variants for proxy-based evaluation:
1) a smaller general-purpose language model \palm-S; and 2) \palm-S$^*$ which is specialized in coding~\citep{anil2023palm}.
Note that while we primarily use proxy-based metrics to evaluate rationales, we also perform human ratings of rationale quality (\cref{sec:human_rating}), with results in line with our proxy-based evaluation.

\textbf{Hyperparameters.}\quad
We perform temperature sampling ($T=0.8$) with a sample size of 32 for training ($|S_j|=32$ in \cref{alg:next-iter}) and \patks{k} evaluation.
In the first iteration in \cref{alg:next-iter}, we use \patks{1} estimated with these 32 samples as the filtering metric $\mathcal{M}(\cdot)$ to find challenging problems whose $\mathcal{M}(\cdot) \leq 10\%$ for training.
We perform $10$ iterations of \ours training and pick the best model using \patks{1} on the development set.

\begin{figure*}[!t]
    \centering
    \includegraphics[width=\linewidth]{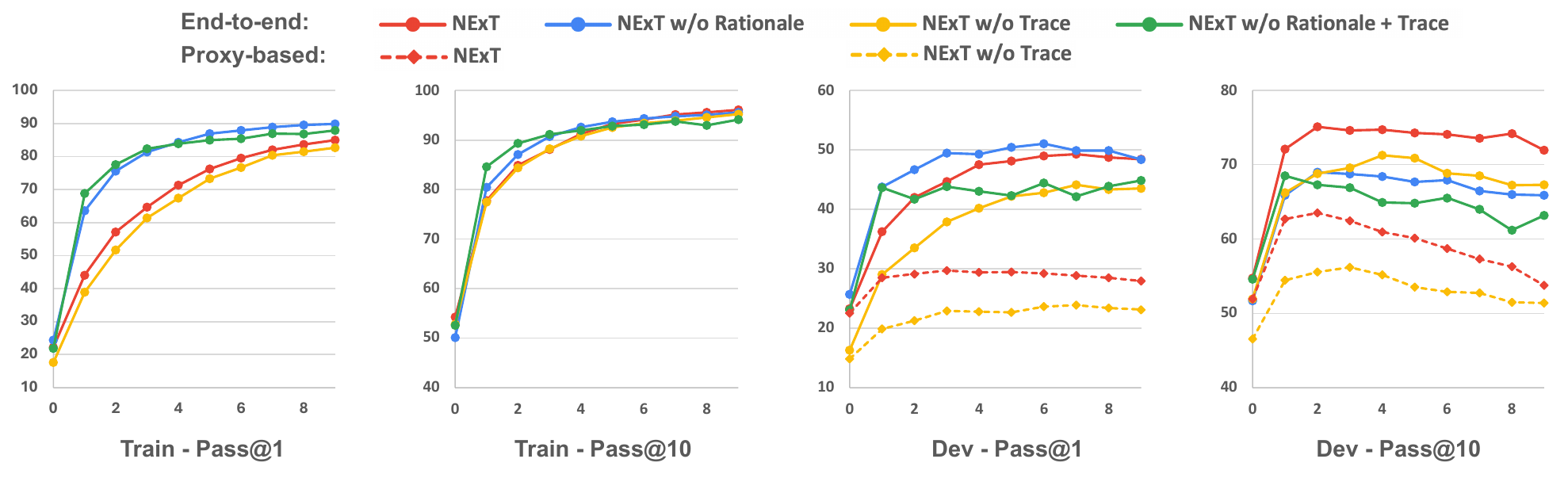}
    \caption{Ablations on removing rationales and/or traces during the iterative training of \ours.
    Note that different min/max values are taken for $y$-axis for clarify among different curves but consistent gridline intervals are used for easier comparison.
    }
    \label{fig:all_traj}
\end{figure*}
\subsection{Main Results}
\label{sec:main_results}
In our experiments, we compare our model with strong LLMs (used in \cref{sec:pre_study}), analyze the impact of rationales and program traces, and perform generalization experiments on \hepack/ and human evaluation of rationale quality.

\textbf{\ours improves program fix rate.}\quad
We first compare the end-to-end program repair performance of \palm-L before and after \ours training (\textbf{\palm-L$+$NExT}) in \cref{tab:compare-gpt} (\textit{Left}).
\ours leads to significant improvements on the end-to-end fix rates across the board, with a $26.1\%$ absolute improvement on \patks{1}.
Interestingly, the gain on \patks{k} is generally higher for smaller $k$.
This might suggest that the model becomes more confident about program fixes after \ours training, while the sample diversity also improves, as indicated by improved \patks{25}.
For reference, we also include results from GPT models.
Notably, \ourpalm outperforms GPT-3.5 on all \patk metrics.

\textbf{\ours improves rationale quality.}\quad
\autoref{tab:compare-gpt} (\textit{Right}) shows the improvements of \ourpalm on our proxy-based evaluation, where we approximate rationale quality using the performance of smaller LMs when conditioned on those rationales.
Again, \ours yields consistent improvements across all \patks{k} metrics.
This suggests that \ours improves \palm-L's skill in reasoning with execution to solve \mbppr/ problems, leading to rationales that are more helpful for smaller LMs.
In \cref{sec:case-study}, we present a case study to demonstrate different reasoning strategies \ourpalm adopts to repair programs using execution information.
As we later show in \cref{sec:human_rating}, our proxy-based metrics are also consistent with human ratings, and rationales from \ourpalm are strongly preferred by annotators compared to those from \palm-L.

\textbf{\ourpalm outperforms strong LLMs.}\quad
We compare \ourpalm with a series of strong LLMs from the preliminary study (\cref{sec:pre_study}) in \autoref{fig:external-comparison}.
\ourpalm outperforms strong open-source LLMs by a minimum of 29.4\% and 11.1\% on end-to-end and proxy-based \patks{1} results, respectively, while on par with GPT-3.5.
These results show that \ourpalm is a competitive model on program repair by reasoning with execution.

\textbf{Learning to reason in natural language improves generalization and sample diversity.}\quad
To further demonstrate the importance of using CoT reasoning in \ours self-training,
we compare \ourpalm with a strong self-training-based program repair model implemented in \ours, which directly generates code fixes using runtime execution information without CoT reasoning.
This ablation resembles SelfAPR \citep{Ye2022SelfAPRSP}, which also adopts self-training to iteratively synthesize data using unit test diagnostics, while our ablation uses traces with richer execution information.
\cref{fig:all_traj} shows model performance w.r.t.~\ours training iterations.
When trained without CoT reasoning (\textbf{\ours w/o rationale}), \palm-L converges much faster on the training set, which is not surprising since the model only learns to generate code fixes without additional reasoning tasks such as explaining bugs in NL.
However, on the \textsc{Dev} set, \ourpalm still outperforms this baseline in \patks{10} with comparable \patks{1} accuracy, and the gap on \patks{10} becomes larger with more iterations.
This shows that by reasoning in natural language, \ourpalm generalizes much better to unseen \mbppr/ problems with greater sample diversity.
In \cref{fig:all-traj} of \cref{sec:app_full_training_dev_results}, we also show that the gain from \ourpalm against this ablation on \patks{k} is even more pronounced for larger $k > 10$, which suggests that learning to reason in CoT rationales improves sample diversity on program repair, similar to the findings on other code generation tasks~\citep{yin2023arcade}.

\textbf{Reasoning with execution traces is critical.}\quad
To understand the importance of leveraging program traces to reason with execution, we compare with an ablation of \ours \emph{without} using program traces, which follows the same procedure in \cref{alg:next-iter} except that traces $\tr$ are not used to generate rationales in Line \ref{alg:next:sampling} (\textbf{\ours w/o traces}, \cref{fig:all_traj}). This variant can also be seen as a direct application of the rationale generation bootstrapping method in \citet{zelikman2022star}, which trains a model on sampled rationales that lead to correct task solutions without relying on additional execution information.
Without traces, \palm-L is consistently worse than \ourpalm on the \textsc{Dev} set across iterations, both in terms of end-to-end fix rate and proxy-based metrics.
This suggests that reasoning with execution information is critical for \palm-L on program repair tasks.
Interestingly, while the gap on the development set is significant, the two models achieve similar scores on the training set, which suggests that reasoning with pre-existing execution traces also help the model generalize better to unseen tasks at test-time.

\textbf{Our model works without traces at test-time.}\quad
While program traces are crucial for reasoning with execution, such execution information may not always be available at test time (\eg when execution is prohibitively expensive).
To stress-test \ourpalm in scenarios where execution information is absent, we remove execution traces from its input at test time in \autoref{tab:test-no-tracing}.
\ourpalm still yields an end-to-end fix rate of $40.8\%$, which is an $21.8\%$ improvement over the 3-shot \palm-L baseline and is only $3.3\%$ lower than \ours trained without traces, for which is tested in-distribution.
The results from the proxy-based evaluation of rationales are also consistent with the fix rate.

\begin{table}[t]
\centering
\small
\setlength{\tabcolsep}{3pt}
\begin{tabular}{@{}lcccc@{}}
\toprule
             & \multicolumn{2}{c}{\textbf{Test w/ Trace}} & \multicolumn{2}{c}{\textbf{Test w/o Trace}} \\\cmidrule{2-5}
\textbf{Methods}                       & E2E                  & Proxy               & E2E                   & Proxy               \\\midrule
\palm-L                & 23.2                 & 22.5                & 19.0                  & 14.8                \\
$+$\ours (w/ trace)  & 49.3$_{+26.1}$       & 28.8$_{+6.3}$       &\cellcolor{red!35}40.8$_{+21.8}$        & \cellcolor{red!35}19.5$_{+4.7}$       \\
$+$\ours w/o trace & $-$                   & $-$                  & 44.1$_{+25.1}$        & 23.9$_{+9.1}$        \\\bottomrule
\end{tabular}
\caption{\ourpalm trained with traces outperforms \palm-L when traces are absent at test time as shown in \colorbox{red!35}{highlighted results}. Results are on \mbppr/; \textbf{Test w/ Trace:} results from \cref{tab:compare-gpt}.}
\label{tab:test-no-tracing}
\vspace{-2mm}
\end{table}

\begin{table}[t]
\centering
\small
\begin{tabular}{lll}
\toprule
\textbf{Models} / \textbf{\patks{1}} & \textbf{End-to-End}   & \textbf{Proxy-based} \\\midrule
    \multicolumn{3}{c}{\textit{Baselines w/ 3-shot prompting}} \\[3pt]
Mistral-7B$^*$               & 12.8           & 16.5           \\
OctoCoder-15.5B$^*$          & 17.7           & 17.7           \\
StarCoder-15.5B$^*$          & 14.6           & 13.1           \\
DeepSeekCoder-33B$^*$        & 28.0           & 18.3           \\
Mixtral-8x7B$^*$            & 32.3           & 30.8           \\
GPT-4                        & 77.6           & 56.6           \\
GPT-3.5                      & 59.4           & 41.8           \\ [0.5mm]\hdashline\noalign{\vskip 1mm}
PaLM-2-L                     & 32.2           & 31.9           \\
PaLM-2-L w/o tracing$^\dagger$ & 30.3           & 30.4           \\\midrule
\ourpalm                     & 42.5$_{+10.3}$ & 38.0$_{+6.1}$  \\
\quad w/o tracing$^\dagger$   & 38.1$_{+7.8}$  & 30.6$_{+0.2}$  \\
\quad w/o rationale           & 44.5$_{+12.3}$ & $-$              \\
\quad w/o tracing + rationale$^\dagger$ & 31.4$_{+1.1}$  & $-$             \\\bottomrule
\end{tabular}
\caption{
Generalization results on \hepack/. \ourpalm models are only trained with \mbppr/.
$^*$obtained using greedy decoding; $^\dagger$no traces provided at test time.}
\label{tab:he_results}
\end{table}

\begin{table}[t]
    \centering
    \setlength{\tabcolsep}{2pt}
    \begin{tabular}{l@{\hspace{2mm}}>{\centering}p{8mm}>{\centering}p{4mm}>{\centering}p{8mm}>{\centering}p{8mm}>{\centering}p{4mm}>{\centering}p{8mm}cc}
        \toprule
        & \multicolumn{3}{c}{\textbf{Explain bugs?}} & \multicolumn{3}{c}{\textbf{Suggest fixes?}} &  & \\ \cmidrule(lr){2-4}\cmidrule(lr){5-7}
        &  \ccmark & \ycmark & \xxmark & \ccmark            & \ycmark            & \xxmark            & \textbf{Overall} & \textbf{Best?} \\ \midrule
        GPT-3.5 & 43 & 26 & 35 & 44 & 16 & 44 & 51.9\% & 34.6\% \\ [0.5mm]\hdashline\noalign{\vskip 1mm}
        \palm-L & 27 & 24 & 53 & 31 & \phantom{0}5 & 68 & 34.9\% & \phantom{0}6.7\% \\ 
        ~~$+$\ours & 48 & 24 & 32 & 42 & \phantom{0}6 & 56 & 50.5\% & 32.7\% \\ \bottomrule 
    \end{tabular}
    \caption{Results for human annotation of rationale quality. Base models use 3-shot prompting. Numbers under the questions are counts of ratings.}
    \label{tab:human_rating}
\end{table}

\textbf{Our model generalizes to \hepack/ at test-time. }\quad
To further evaluate the generalization ability of \ourpalm, we test our model (trained on \mbppr/) on \hepack/.
\cref{tab:he_results} summarizes the results.
\ours achieves reasonable generalization on \hepack/, outperforming the base \palm-L model by a large margin (\ie $14.3\%$ on end-to-end fix rate and $6.0\%$ on proxy evaluation). 
Aligned with our previous findings on \mbppr/ in \cref{fig:all_traj}, reasoning with execution traces (\textit{c.f.}~w/o traces) improves fix rate and rationale quality.
Moreover, we remark that with iterative learning, \ourpalm is on par with the strong program repair method without CoT reasoning (w/o rationale), similar to the results on \mbppr/.
This is in contrast with our preliminary study in \cref{sec:pre_study}, where \palm-L with CoT prompting is much worse than vanilla prompting without using rationales.
Overall, these results indicate that \ourpalm could robustly generalize to out-of-distribution repair tasks without additional dataset-specific finetuning.

\subsection{Human Evaluation of Rationale Quality}
\label{sec:human_rating}

Our proxy-based evaluation suggests the extrinsic value of the CoT rationales from \ourpalm.
We further conduct an intrinsic evaluation by manually rating the quality of model-predicted rationales on 104 sampled \mbppr/ repair tasks from the \textsc{Dev} set. 
Specifically, we ask raters to judge the quality of rationales generated by three models (\ourpalm, \palm-L and GPT-3.5) in a three-way side-by-side setting.
Each rationale is rated in two aspects:
(1) its helpfulness in explaining bugs ($Q_1$, \eg first two paragraphs in $\hr$, \cref{fig:next}),
and (2) its helpfulness in suggesting code fixes ($Q_2$, \eg \goodratspan{\textit{``a fixed version that uses $\ldots$''}} in $\hr$).
Each question has a three-scale answer (\ccmark~Completely correct and very helpful;
\ycmark~Partially correct with minor errors but still helpful; \xxmark~Incorrect and not helpful).
We also compute an \textbf{overall score} of rationale quality using numeric values of $\{+1, 0.5, 0\}$ for the three scales and averaged over $Q_1$ and $Q_2$.
Finally, we ask raters to pick a single \textbf{best choice} if there is not a clear tie.
More details about our human evaluation pipeline is described in \cref{sec:annotation-details}.

\cref{tab:human_rating} summarizes the result.
Compared to the base \palm model, \ourpalm generates significantly more high-quality rationales with correct explanations of bugs and fix suggestions.
Additionally, compared to GPT-3.5, \ourpalm also has more rationales with correct bug explanations, while interestingly, GPT-3.5 generates more rationales with partially correct fix suggestions.
We hypothesize that including more exemplars with detailed fix suggestions to our few-shot prompts during \ours training (\cref{sec:full-prompt}) would help mitigate this issue.
Nevertheless, the overall scores and rater-assigned best choice suggest that the rationales predicted by \ourpalm are of significantly higher quality compared to those from \palm-L, and are on par with the predictions from GPT-3.5.
Overall, this finding is in line with the proxy evaluation results in \cref{fig:external-comparison} (GPT 3.5 $\approx$ \ourpalm $\gg$ \palm-L), suggesting that the latter is a reasonable surrogate metric for rationale quality.
In \cref{sec:case-study}, we present example generated rationales that show a variety of reasoning patterns.

\section{Related Work}
\label{sec:related-work}

\textbf{Reasoning about Program Execution}\quad
Several lines of research has explored learning methods to reason about program execution.
Program synthesis systems often leverage the execution states of partially generated programs~\citep{shin18synthesis,Wang2018RobustTG,Chen2021LatentEF,shi2022crossbeam} or the next execution subgoals~\citep{shi2024exedec} to guide search in sequence-to-sequence models.
There has also been work on training neural networks to mimic program
execution, like a learned interpreter \citep{Zaremba2014LearningTE,Bieber2020-ot,nye2021show}, often with specialized neural architectures to model the data flow of program execution~\citep{Graves2014NeuralTM,Gaunt2016DifferentiablePW,Bosnjak2016ProgrammingWA,Bieber2022StaticPO}.
Instead of using domain-specific architectures to encode and reason about program execution, our work focuses on teaching LLMs to reason with execution in natural language.
In particular, \textit{Scratchpad}~\citep{nye2021show} and \textit{Self-Debugging}~\citep{chen2023teaching} are two notable works that also models execution traces using LLMs.
The core difference is that these methods focus on predicting reasoning chains that contain trace information, such as executed lines with variable states~\citep{nye2021show} or their natural language summaries~\citep{chen2023teaching}.
On the other hand, \ours aims to leverage existing execution traces from a runtime to aid the reasoning process, which often leads to more compact rationales tailored for downstream tasks.
We present a more detailed comparison and discussion on \ours and these related works in \cref{sec:comp-sp-sd}.

\textbf{Program Repair}\quad
Several works in program repair have leveraged execution information such as traces~\citep{Gupta2020SynthesizeEA,Bouzenia2023TraceFixerET} or test diagnostics~\citep{Xia2023KeepTC,Ye2022SelfAPRSP}.
Different from \citet{Bouzenia2023TraceFixerET} which represents traces by directly pairing unrolled executed lines with their variable states, \ours inlines indexed variable states as code comments, which is more token efficient while preserving the original code structure.
Similar to \ours, \citet{Ye2022SelfAPRSP} construct synthetic self-training data using test execution results, while our approach generates both NL rationales and fixed programs with better interpretability.
Recently, LLMs have been applied to program repair~\citep{Fan2022AutomatedRO,Xia2022LessTM,Xia2023AutomatedPR,Sobania2023AnAO,Paul2023EnhancingAP,Jiang2023ImpactOC}.
Among them, \citet{Kang2023ExplainableAD} uses a ReAct-style CoT reasoning loop~\citep{Yao2022ReActSR} to predict repair actions based on interactive  feedback from debuggers, while \ours focuses on tuning LLMs to reason with pre-existing execution information  without intermediate feedback.
Finally, as a related stream of research, self-improvement methods iteratively refine a model's code solutions using CoT reasoning over self-provided~\citep{Madaan2023SelfRefineIR} or test-driven feedback~\citep{chen2023teaching,olausson2023demystifying}.
Instead of relying on high-level execution signals like error messages, \ours trains LLMs to reason with step-wise program traces.
Our learnable rationales are also more flexible without following a predefined reasoning template.
Besides, since traces already capture rich execution semantics, the resulting rationales could be more succinct and targeted to the downstream task (\eg explain bugs), without redundant reasoning steps to trace the program by the model itself to recover useful execution information.

\textbf{Supervised CoT Reasoning}\quad
LLMs can solve problems more accurately when instructed to work out the answer step by step in a \emph{chain of thought} or a \emph{scratchpad} \citep{chainofthought,nye2021show,rajani2019explain,shwartz2020selftalk}.
Improvements on this approach involve
finetuning LLMs on chain-of-thought reasoning data.
Such CoT data is either manually curated \citep{Chung2022ScalingIL,Longpre2023TheFC,lightman2023let}, or distilled from more capable teacher models~\citep{Gunasekar2023TextbooksAA,Mukherjee2023OrcaPL,Mitra2023Orca2T,Fu2023SpecializingSL}.
Instead of relying on labeled or distilled data, \ours uses self-training to iteratively bootstrap a synthetic dataset of high-quality rationales with minimal manual annotation. Our work differs from previous work using bootstrapping
~\citep{zelikman2022star,Hoffman2023trice} 
in the type of rationales and the use of execution information; see~\cref{sec:methodology} for more discussion.
While we use the correctness of the program fix for filtering the rationales, which is reminiscent of outcome supervision; it is also possible to use process supervision with human annotations~\citep{Uesato2022SolvingMW, lightman2023let}, or obtain such supervision automatically by estimating the quality of each step using Monte Carlo Tree Search~\citep{wang2024mathshepherd} and by identifying partially-correct program prefixes~\citep{ni2022learning}. 
Finally, existing research has investigated finetuning of LLMs to predict the execution information directly, such as predicting line-by-line execution traces~\citep{nye2021show}, abstract runtime properties~\citep{Pei2023CanLL}, or final output~\citep{Zaremba2014LearningTE,Bieber2020-ot}. \ours addresses a different problem; instead of predicting the execution information, \ours takes it as given, and instead learns to discover flexible task-specific NL rationales that aid a downstream programming task.

\section{Conclusion}
In this paper we present \ours, a self-training method to finetune LLMs to reason with program execution given traces.
We demonstrate that \palm-L trained using \ours yields high-quality natural language rationales and achieves stronger success rates on two program repair tasks.
As future work, we plan to apply \ours to a broader range of program understanding tasks while expanding the trace representation to support more programming languages.

\section*{Acknowledgements}

We would like to express our sincere gratitude to Martín Abadi, Xinyun Chen, Hanjun Dai, Kexin Pei and members of the Learning for Code team at Google DeepMind for their invaluable feedback.
We are also grateful to Austin Tarango for his support to this work.

\bibliography{main}

\appendix
\onecolumn

\section{Additional Details of \ours}
\subsection{Details for Inline Trace Representation}
\label{sec:trace-repr-details}

\paragraph{Definitions.} A program $y\in\mathcal{Y}$ consists of a sequence of statements $\{u_1,...,u_m\}$. And a program state $h$ is a mapping between identifiers (\ie variable names) to values, \ie $h \in \{k\mapsto v| k\in\mathcal{K}, v\in\mathcal{V}\}$.
Given an input to the program, an execution trace is defined as a sequence of program states, \ie $\tr = \{h_1,...,h_t\}$, which are the results after executing the statements with the \textit{order of execution}, \ie $\{u_{e_1}, u_{e_2},...,u_{e_t}\}$. In this way, the relation between program statements and execution states can be seen as a function that maps from states to statements, \ie $h_i \mapsto u_{e_i}$, because each statement could be executed multiple times due to loops or recursion.

\paragraph{Program state representation.}
For typical programs, most of the variable values will stay the same between two adjacent states $h_{i-1}$ and $h_i$. Thus to save tokens, we represent a state $h_i$ only by the variables that have changed the value compared with the previous state $h_{i-1}$. And we use a reified variable state representation, \ie using the grammar for an init function in Python (\eg \code{lst=[1, 2, 3]}). Note that it is possible for a statement to have no effect on any traceable variables (\eg ``\code{pass}'', or ``\code{print}'', or ``\code{lst[i]=lst[i]}''). 
To distinguish this case with unreached statements (\eg ``\code{else}'' branch that next got executed), we append a string ``\code{NO\_CHANGE}'' instead.
In addition to the variable state, we number all the states by the order of execution and prepend the ordinal number to the beginning of the state, \eg ``\code{(1) odd\_list=[]}'' in \cref{fig:trace-example}. 

\paragraph{Inline trace representation.}
To obtain the inline trace representation, we first group the program states in a trace $\tr$ by the corresponding program statements to collect a sequence of states for the same statement $u_i$ as $H_i=\{h_j|u_{e_j} = u_i\}$, and we order the states in $H_i$ by the execution order.
For statements inside a loop body, or a function that is called recursively, the number of corresponding states can be very large. In order to further save tokens, if $|H_i|>3$, we will only incorporate the first two states and the last state, and skip the ones in the middle.
After that, we simply concatenate all the state representations with the semicolon ``\code{;}'' as the delimiter, and append it after the statement itself $u_i$ following a hash ``\code{\#}'' to note it as an inline comment.
An example of the resulting representation is ``\code{even\_list.append(n)  \# (4) even\_list=[1]; (8) even\_list=[1, 3]; ...; (20) even\_list=[1, 3, 5, 7, 9];}'', as shown in \cref{fig:trace-example}.

\paragraph{Limitations.}
First of all, our tracing framework currently do not extend beyond native Python programs, thus it can not trace code that is not written in Python (\eg C code in \code{numpy}).
One other limitation of our tracing representation is that for ``\code{if}'' conditions, though it would be better to leave traces of ``\code{(1) True; (2) True; (3); False;}'', currently our tracing framework that based on the ``\code{sys.settrace()}'' hook of Python does not capture this. However, since we labeled all the states by the execution order, the LLMs can infer the conditions by the fact that certain branch is taken.
Another limitation is the representation of Collections. Currently we still present all the elements in a collection, and empirically it works well with benchmarks as \mbppr/ and \hepack/.
However, certain heuristics may be needed to skip certain elements (\eg like the one we use to skip certain states in a loop) to be more token efficient.
For more complex objects (\eg Tensors, DataFrames), while we can define heuristics to represent key properties of those objects in traces (\eg ``a float tensor of shape 128 x 64'',``a Dataframe with columns Name, Math, ...''), perhaps a more interesting idea would be to let the models decide which properties they would inspect and generate relevant code (\eg ``\code{tensor.shape}'' or ``\code{df.head(3)}'') to inspect them in a debugger or interpreter (\eg pdb). 
The same idea can be applied to longer programs, as the model can selectively decide which lines of code to inspect and create traces for, similar to how human developers debug programs. We will leave these as exciting future directions.

\subsection{Details for Iterative Self-Training}
\paragraph{Bootstrapping rationales and fixes via temperature sampling.}
To avoid potential ``cold start'' problem \citep{liang2018memory, ni2020merging}, for the first iteration, we use few-shot prompting with three exemplars (shown in \cref{sec:full-prompt}) and set the sample size to 96. For all later iterations, we use zero-shot prompting as the model is already adapted to the style of the rationales and fixes after the first round of finetuning, and we set the sample size to 32. We set the sampling temperature $T=0.8$ for all iterations.

\paragraph{Filtering rationales and fixes.}
Given the inputs in the prompt, we sample the rationale and fixes in tandem. To separate the natural language rationale and the program fix, we use an regular expression in Python to extract the content between two  sets of three backticks (\lstinline|```|), which is commonly used to note code blocks in markdown.\footnote{For the strong LLMs that we used in this work, we did not observe any issue for following this style, which is specified in the few-shot prompt. The only exceptions are with GPT models, where they typically append the language (\ie ``python'') after the first set of backticks (\eg \lstinline|```|python), which we also handled with \code{regex}.}
After we filter out the rationales and fixes that are incorrect using the test cases, we create the training set by sub-sampling correct ``(rationale, fix)'' pairs to allow a maximum of 3 correct fixes with their rationales for each problem in \mbppr/. This is to balance the number of rationales and fixes for each problem and avoid examples from certain examples (typically easier ones) being overly represented in the training set.

\subsection{Discussion with Previous Work}
\label{sec:comp-sp-sd}
Here we discuss \ours in the context of two important previous work in the domain of reasoning about program execution, namely \textit{Scratchpad}~\citep{nye2021show} and \textit{Self-Debugging}~\cite{chen2023teaching}. Such comparison is also characterized by \cref{tab:scratchpad-self-debug-comp}.

\begin{table}[t]
    \centering
    \begin{tabular}{lccc}
    \toprule
        \textbf{Methods}        & Use of Trace  & Rationale Format & Model Fine-tuning \\\midrule
        \ours                   & Input         & Natural Language & Yes \\
        Scratchpad~\citep{nye2021show}              & Output        & Scratchpad Repr.  & Yes \\
        Self-Debugging~\citep{chen2023teaching}          & Output        & Natural Language & No  \\\bottomrule
    \end{tabular}
    \caption{Comparison between the methods proposed in \ours, Scratchpad, and Self-Debugging.}
    \label{tab:scratchpad-self-debug-comp}
\end{table}

\begin{table}[t]
    \centering
    \begin{tabular}{lcccccccc}
    \toprule
        \multirow{2}{*}{\textbf{Trace Repr.}} & \multicolumn{8}{c}{\textbf{Length Cutoff (\# Tokens)}} \\\cline{2-9}
                            & 128   & 256   & 512     & 1,024    & 2,048   & 4,096     & 8,192      & 16,384  \\\midrule
        Inline (ours)       & 0.1\% & 7.3\% & 37.5\%  & 78.9\%  & 95.1\% & 98.5\%   & 99.2\%    & 99.5\% \\
        Scratchpad          & 0.0\% & 0.2\% & 15.1\%  & 38.2\%  & 60.1\% & 76.1\%   & 85.1\%    & 92.1\% \\\bottomrule
    \end{tabular}
    \caption{Percentage of \mbppr/ examples that can be fit into different context windows using different trace representations (\ie ours and \citet{nye2021show}). Traces of all three tests are included.}
    \label{tab:trace_repr_length}
\end{table}

\paragraph{\textit{Scratchpad} and \ours.} Similarly to \ours, \citet{nye2021show} also proposed to use execution traces to help the LLMs to reason about program execution. However, \citet{nye2021show} aimed to generate these traces as intermediate reasoning steps at inference time, either via few-shot prompting or model fine-tuning. Yet in \ours, we use execution traces as part of the input to the LLMs, so they can directly use the execution states to ground the generated natural language rationales. Moreover, we choose to use natural language as the primary format for reasoning, which is more flexible and easier to be understood by the human programmers.
We also perform a length comparison of our proposed inline trace representation with the scratchpad representation proposed in \cref{tab:trace_repr_length}, and results show that our proposed inline trace representation is much more compact than scratchpad.

\paragraph{\textit{Self-Debugging} and \ours.} Self-Debugging~\citep{chen2023teaching} is a seminal approach that also performs CoT reasoning over program execution to identify errors in code solutions. Different from \ours, Self-Debugging can optionally leverages high-level execution error messages to bootstrap CoT reasoning, while our method trains LLMs to reason with concrete step-wise execution traces.
In addition, Self-Debugging also introduced a particular form of CoT rationales that resemble step-by-step traces in natural language. 
Notably, such rationales are generated by LLMs to aid the model in locating bugs by simulating execution in a step-by-step fashion. They are not the ground-truth execution traces generated by actually running the program. As we discussed in \cref{sec:related-work}, in contrast, our model relies on existing traces from program execution. Since those traces already capture rich execution information, intuitively, the resulting CoT rationales in \ours could be more succinct and ``to the point'' without redundant reasoning steps to ``trace'' the program step-by-step by the model itself in order to recover useful execution information. 

Finally, we remark that our ``Test w/o Trace'' setting in \cref{sec:main_results} shares similar spirits with the setup in Self-Debugging, as both methods perform CoT reasoning about execution without gold execution traces. From the results in \cref{tab:test-no-tracing}, \ours also greatly improves the model's ability to repair programs even without using gold execution traces at test time. This may suggest that \ours can potentially improve the self-debugging skills of LLMs through iterative training, for which we leave as exciting future work to explore.

\section{Experiment Setup Details}

\subsection{Creating \mbppr/}
\label{sec:mbppr-details}
The original \mbpp/ dataset \cite{austin2021program} consists of three splits, \ie train/dev/test sets of 374/90/500 Python programming problems. 
To increase the number of training example, we first perform a re-split of the original \mbpp/ dataset, by moving half of the test data into the training split, resulting in 624/90/250 problems in the re-split dataset.
Then for each \mbpp/ problem in the re-split train and dev set, we collect a set of failed solutions from the released model outputs in \citet{ni2023lever}.
More specifically, we take the $100$ samples for each problems, filter out those correct solutions, and keep the ones that do not pass all the tests. As different problems have various number of buggy solutions, we balance this out by keeping at most $20$ buggy solutions for each \mbpp/ problem.\footnote{This actually biased the dataset towards harder problems as easier problems may not have more than 20 buggy solutions from 100 samples, thus it might be one of the reasons for repairing solutions in \mbppr/ to be more challenging than generating code for the original \mbpp/ dataset.}
This yields the \mbppr/ dataset, with $10,047$ repair tasks in the training set and $1,468$ examples in the dev set.

\subsection{Use of test cases.}
\label{sec:use-test-case}
For each program repair task, there is typically a set of open test cases that are used for debugging purposes, as well as a set of hidden test cases that are only used for evaluation of correctness.
When we generate traces using test cases, we use only the open test cases and only feed the open test cases to the model as part of the prompt. Then when we evaluate the generated fix, we resort to all test cases (\ie open + hidden tests) and only regard a fix as correct when it passes all test cases.
While the \he/ dataset makes this distinction between open and test cases, the \mbpp/ dataset does not make such distinction. Thus for \mbppr/, we use all test cases both as the inputs and during evaluation. 
While this may lead to false positives when the fixes are overfit to the test cases, and we did find such case during human annotations. 

\begin{figure}[!h]
    \centering
    \fbox{\includegraphics[width=0.8\linewidth]{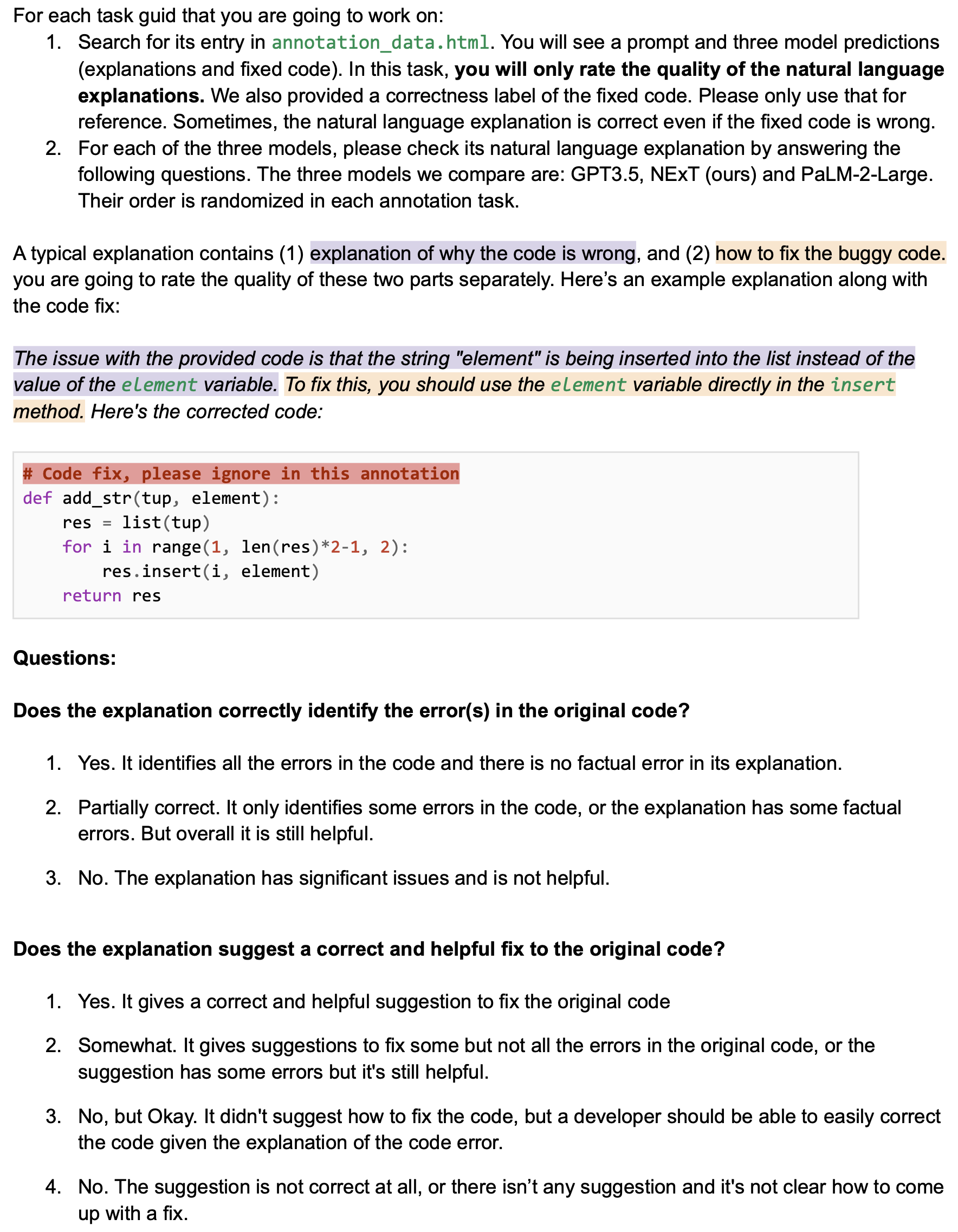}}
    \caption{Instructions for the human annotators when annotating the quality of the model generated rationales.}
    \label{fig:annotation-inst}
\end{figure}

\subsection{Details of Human Annotation of Rationale Quality}
\label{sec:annotation-details}
We annotated model predictions on 104 sampled \mbppr/ repair tasks from the \textsc{Dev} set.
Those fix tasks are randomly sampled while ensuring that they cover all the 90 dev \mbpp/ problems.
All the tasks are pre-screened to be valid program repair problems.
Annotation is performed in a three-way side-by-side setting.
Models are anonymized and their order is randomized.
Raters are asked to judge the quality of rationales from three models (\ourpalm, \palm-L and GPT-3.5) on the same \mbppr/ problem.
Each rationale is rated from two aspects: 
(1) its helpfulness in explaining bugs ($Q_1:$ \textit{Does the rationale correctly explain bugs in the original code?} \eg first two paragraphs in $\hr$, \cref{fig:next}),
and (2) its helpfulness in suggesting code fixes ($Q_2:$ \textit{Does the rationale suggest a correct and helpful fix?} \eg \goodratspan{\textit{``a fixed version that uses $\ldots$''}} in $\hr$, \cref{fig:next}).\footnote{We only rate the quality of rationales (not the fixed code), while we still show the predicted fixed code to raters for reference.}
Each question has a three-scale answer (\ccmark~Completely correct and very helpful
; \ycmark~Partially correct with minor errors but still helpful; \xxmark~Incorrect and not helpful).
In a pilot study, we find that fix suggestions could often be redundant if the rationale already contains detailed explanation of bugs such that a developer could easily correct the code without an explicit fix suggestion (\eg Example 2, \cref{sec:case-study}).
Therefore, for $Q_2$, we also consider such cases as correct (\ccmark) if a model didn't suggest a fix in its rationale but the fix is obvious after bug explanations.
We list our annotation guideline in \cref{fig:annotation-inst}. Note that for $Q_2$, both answers {\tt (1)} and {\tt (3)} are counted as correct (\ccmark) answers.

\section{Additional Experiment Results}
\label{sec:app_full_training_dev_results}
\begin{figure*}[!t]
    \centering
    \includegraphics[width=\linewidth]{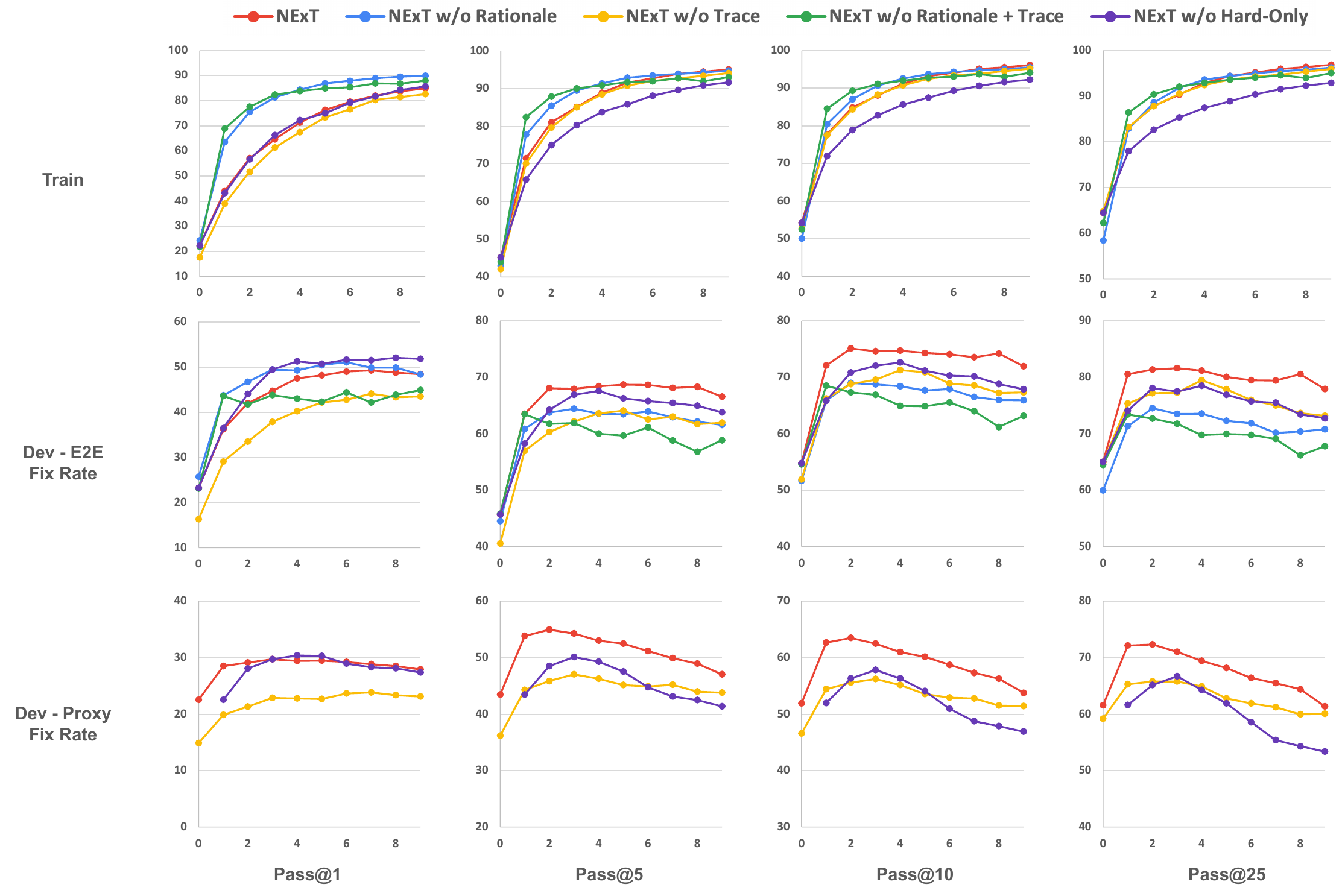}
    \caption{\patk performance on the train and dev sets of \mbppr/ for \ours and all its ablations.}
    \label{fig:all-traj}
\end{figure*}

Here we show the learning curve of \ours and all its ablations in \cref{fig:all-traj}. We also show the full results for \mbppr/ and \hepack/ in \cref{tab:full-mbppr-results} and \cref{tab:full-hepack-results}, respectively. 

\paragraph{Learning CoT rationales further improves \patks{25}.}
From \cref{sec:main_results}, we mention that learning to reason in natural language improves sample diversity, registering higher \patks{10} than the baseline of finetuning for generating fixes only (\textbf{\ours w/o Rationale)}. From \cref{tab:full-mbppr-results} and \cref{tab:full-hepack-results}, we can observe that such performance advantage is even larger with \patks{25}, with 7.6\% improvements on \mbppr/ and 6.8\% improvements on \hepack/.

\paragraph{Training on hard-only examples.}
One part of our data filtering pipeline is to only perform sampling and train on the samples from hard problems (\cref{sec:methodology}).
Here we discuss more about the benefits and potential issues of doing so, by presenting results on a ``\textit{w/o hard-only}'' ablation, where the model learns from rationales and fixes from both hard and easy examples.
Efficiency-wise, by only sampling on the hard example, which is around half of the problems, we greatly can accelerate the sampling process.
And from results in \cref{fig:all-traj}, only training with hard example also comes with performance benefits under the iterative self-training framework. More specifically, we notice a non-trivial gap between the training curve of this ``w/o hard-only'' baseline and the rest of the ablations, especially for \patks{10} and \patks{25} performance on the training set.
This means that the model trained on both easy and hard examples leads to more problems in the training set unsolved (\ie none of the samples are correct), and no learning signal can come from such problems.
This also reflects on the dev set performance. While it is worth noticing that the end-to-end \patks{1} performance for ``w/o hard-only'' is slightly better than \ours trained only trained on hard examples, it performs worse in all other evaluations, with the trend of larger gaps with higher $k$ values for \patk, especially for the proxy-based evaluation.
This suggests that training on hard examples not only improves sample efficiency, but also improves the general fix rate as well as the quality of the generated rationales.

\paragraph{Proxy-based evaluation results are consistent with different proxy models.}
In the previous proxy-based evaluation \cref{sec:main_results}, we report the proxy-based fix rates by averaging over the performance using \palm-S and \palm-S$^*$ as the proxy models. In \cref{tab:full-mbppr-results} and \cref{tab:full-hepack-results}, we show the separated results for different proxy models. From these results, we can observe that the relative rationale quality evaluated by different proxy models are largely consistent, with the stronger proxy model (\palm-S$^*$) having better proxy-based fix rates. 
In addition to the consistency we show with human annotations, this shows the robustness of our proposed proxy-based evaluation method for measuring CoT rationale quality.

\begin{table}[t]
\centering
\scriptsize
\resizebox{\linewidth}{!}{
\begin{tabular}{l|c:cccc|c:cccc|c:cccc}
\toprule
\multirow{3}{*}{\textbf{Models}} & \multicolumn{5}{c}{\textbf{End-to-End Fix Rate}}              & \multicolumn{5}{c}{\textbf{Proxy-based Fix Rate (\palm--S)}} & \multicolumn{5}{c}{\textbf{Proxy-based Fix Rate (\palm--S$^*$)}} \\
                                 & \multicolumn{1}{c}{GD}   & \multicolumn{4}{c}{\patk w/ Sampling} & \multicolumn{1}{c}{GD}     & \multicolumn{4}{c}{\patk w/ Sampling}      & \multicolumn{1}{c}{GD}     & \multicolumn{4}{c}{\patk w/ Sampling}      \\
                                 & \multicolumn{1}{c}{Acc.} & $k$=1  & $k$=5  & $k$=10  & $k$=25 & \multicolumn{1}{c}{Acc.}   & $k$=1   & $k$=5   & $k$=10   & $k$=25   & \multicolumn{1}{c}{Acc.}   & $k$=1   & $k$=5   & $k$=10   & $k$=25   \\\midrule
GPT--3.5                          & 46.4                     & 42.9   & 65.0   & 70.7    & 76.7   & 27.9                       & 24.7    & 46.1    & 54.5     & 64.6     & 31.8                       & 28.5    & 51.5    & 59.5     & 68.2     \\
GPT--3.5 w/o trace                & 47.1                     & 46.8   & 65.9   & 70.7    & 75.7   & 27.2                       & 25.6    & 47.0    & 55.5     & 64.7     & 30.9                       & 30.2    & 53.0    & 60.7     & 68.8     \\
GPT--4                            & 62.8                     & 63.2   & 75.1   & 78.5    & 82.7   & 41.8                       & 42.2    & 64.5    & 71.0     & 76.6     & 47.8                       & 47.4    & 68.5    & 73.9     & 79.0     \\
GPT--4 w/o trace                  & 51.3                     & 44.8   & 68.5   & 73.4    & 78.5   & 29.4                       & 27.1    & 54.2    & 63.4     & 72.2     & 34.9                       & 32.0    & 60.3    & 68.5     & 75.7     \\
\palm-L                         & 26.6                     & 23.2   & 45.7   & 54.7    & 65.0   & 21.5                       & 21.1    & 41.1    & 49.5     & 59.2     & 24.9                       & 23.9    & 45.8    & 54.3     & 63.8     \\
\palm-L w/o trace               & 19.0                     & 16.3   & 42.1   & 52.8    & 64.8   & 14.7                       & 13.7    & 33.9    & 44.1     & 56.7     & 17.4                       & 15.9    & 38.3    & 48.9     & 61.6     \\
\palm-L w/o rationale           & 27.5                     & 25.7   & 44.5   & 51.7    & 60.0   & --                          & --       & --       & --        & --        & --                          & --       & --       & --        & --        \\
\palm-L w/o rationale + trace   & 23.8                     & 23.1   & 45.8   & 54.6    & 64.5   & --                          & --       & --       & --        & --        & --                          & --       & --       & --        & --        \\\midrule
\ours                             & 50.5                     & 49.3   & 68.1   & 73.5    & 79.4   & 25.3                       & 26.1    & 46.8    & 54.4     & 62.9     & 31.8                       & 31.6    & 53.0    & 60.2     & 68.1     \\
\quad test w/o trace             & 41.1                     & 40.8   & 61.8   & 68.9    & 76.4   & 17.6                       & 17.5    & 35.6    & 43.5     & 53.4     & 21.0                       & 21.5    & 42.2    & 50.6     & 60.1     \\
\ours w/o hard--only               & 52.9                     & 52.1   & 65.0   & 68.8    & 73.4   & 23.5                       & 25.1    & 38.6    & 44.0     & 50.9     & 30.0                       & 29.7    & 44.1    & 49.7     & 55.9     \\
\quad test w/o trace             & 41.9                     & 42.2   & 58.1   & 63.2    & 69.2   & 16.3                       & 17.8    & 32.1    & 37.9     & 45.0     & 18.7                       & 21.0    & 36.7    & 43.0     & 50.5     \\
\ours w/o rationale               & 51.8                     & 51.1   & 63.9   & 67.9    & 71.8   & --                          & --       & --       & --        & --        & --                          & --       & --       & --        & --        \\
\quad test w/o trace             & 43.7                     & 43.0   & 57.2   & 61.7    & 66.3   & --                          & --       & --       & --        & --        & --                          & --       & --       & --        & --        \\
\ours w/o trace                   & 44.5                     & 44.1   & 63.0   & 68.5    & 75.0   & 22.3                       & 21.8    & 42.3    & 50.1     & 59.2     & 25.9                       & 25.9    & 48.0    & 55.4     & 63.2     \\
\ours w/o rationale w/o trace     & 46.3                     & 44.9   & 58.9   & 63.2    & 67.8   & --                          & --       & --       & --        & --        & --                          & --       & --       & --        & --        \\\bottomrule
\end{tabular}}
\caption{Full results on \mbppr/. ``GD Acc.'' denotes \patks{1} evaluated with greedy decoding. All models in the top half are few-shot prompted while the bottom half shows the result of \ours and its ablations.}
\label{tab:full-mbppr-results}
\end{table}

\begin{table}[t]
\centering
\scriptsize
\resizebox{\linewidth}{!}{
\begin{tabular}{l|c:cccc|c:cccc|c:cccc}
\toprule
\multirow{3}{*}{\textbf{Models}} & \multicolumn{5}{c}{\textbf{End-to-End Fix Rate}}              & \multicolumn{5}{c}{\textbf{Proxy-based Fix Rate (\palm--S)}} & \multicolumn{5}{c}{\textbf{Proxy-based Fix Rate (\palm--S$^*$)}} \\
                                 & \multicolumn{1}{c}{GD}   & \multicolumn{4}{c}{\patk w/ Sampling} & \multicolumn{1}{c}{GD}     & \multicolumn{4}{c}{\patk w/ Sampling}      & \multicolumn{1}{c}{GD}     & \multicolumn{4}{c}{\patk w/ Sampling}      \\
                                 & \multicolumn{1}{c}{Acc.} & $k$=1  & $k$=5  & $k$=10  & $k$=25 & \multicolumn{1}{c}{Acc.}   & $k$=1   & $k$=5   & $k$=10   & $k$=25   & \multicolumn{1}{c}{Acc.}   & $k$=1   & $k$=5   & $k$=10   & $k$=25   \\\midrule
GPT-3.5                       & 68.9 & 59.4 & 84.5 & 89.2 & 93.0 & 42.1 & 39.0 & 66.1 & 73.4 & 80.2 & 46.3 & 44.6 & 71.6 & 78.8 & 86.8 \\
GPT-3.5 w/o trace             & 65.2 & 65.4 & 85.3 & 89.2 & 92.6 & 45.7 & 41.7 & 68.2 & 76.3 & 84.5 & 50.0 & 47.2 & 73.8 & 81.1 & 88.6 \\
GPT-4                         & 79.9 & 77.6 & 89.3 & 91.1 & 92.9 & 56.1 & 55.4 & 75.7 & 80.8 & 85.8 & 61.0 & 57.7 & 77.5 & 82.7 & 87.4 \\
GPT-4 w/o trace               & 79.3 & 68.9 & 88.3 & 90.7 & 92.9 & 54.9 & 46.1 & 72.3 & 79.0 & 86.1 & 59.8 & 48.7 & 74.4 & 80.8 & 87.5 \\
\palm-L                       & 43.3 & 32.2 & 64.3 & 73.8 & 81.5 & 32.9 & 28.9 & 59.0 & 69.2 & 79.1 & 43.3 & 34.9 & 65.8 & 74.3 & 82.9 \\
\palm-L w/o trace             & 38.4 & 30.3 & 61.9 & 72.9 & 83.3 & 25.6 & 27.8 & 56.2 & 66.0 & 76.6 & 31.1 & 33.0 & 63.5 & 72.7 & 81.8 \\
\palm-L w/o rationale         & 53.0 & 45.3 & 71.5 & 78.9 & 85.4 & --   & --   & --   & --   & --   & --   & --   & --   & --   & --   \\
\palm-L w/o rationale + trace & 48.2 & 43.2 & 71.4 & 80.0 & 87.7 & --   & --   & --   & --   & --   & --   & --   & --   & --   & --   \\ \midrule
\ours                         & 46.3 & 42.5 & 62.6 & 69.1 & 76.5 & 31.7 & 34.8 & 54.8 & 62.4 & 70.2 & 40.9 & 41.3 & 61.8 & 68.9 & 76.4 \\
\quad test w/o trace          & 42.7 & 41.2 & 62.9 & 70.6 & 79.5 & 26.8 & 26.4 & 48.0 & 56.1 & 64.2 & 36.0 & 32.6 & 55.7 & 64.4 & 72.8 \\
\ours w/o hard-only           & 48.8 & 47.7 & 64.8 & 70.4 & 76.6 & 32.9 & 37.2 & 50.8 & 55.5 & 61.9 & 41.5 & 42.4 & 56.3 & 60.8 & 66.9 \\
\quad test w/o trace          & 47.6 & 44.2 & 64.4 & 70.4 & 75.5 & 31.7 & 33.3 & 46.9 & 51.4 & 57.3 & 38.4 & 38.5 & 54.6 & 59.2 & 63.9 \\
\ours w/o rationale           & 47.6 & 44.5 & 58.9 & 63.7 & 69.4 & --   & --   & --   & --   & --   & --   & --   & --   & --   & --   \\
\quad test w/o trace          & 46.3 & 44.7 & 60.4 & 65.2 & 70.2 & --   & --   & --   & --   & --   & --   & --   & --   & --   & --   \\
\ours w/o trace               & 40.9 & 38.1 & 59.1 & 65.3 & 71.5 & 29.3 & 26.9 & 52.1 & 61.1 & 71.5 & 33.5 & 34.4 & 63.1 & 70.8 & 77.4 \\
\ours w/o rationale w/o trace & 30.5 & 31.4 & 44.6 & 49.0 & 54.1 & --   & --   & --   & --   & --   & --   & --   & --   & --   & --   \\\bottomrule
\end{tabular}
}
\caption{Full results on \hepack/. Same notations from \cref{tab:full-mbppr-results} apply.}
\label{tab:full-hepack-results}
\end{table}

\section{Case Study}
\label{sec:case-study}

In this section we present a set of examples to showcase how \ourpalm reasons with program execution to solve \mbppr/ problems. We discover several reasoning patterns the model exhibits that leverage trace information to identify and explain bugs in programs.
First, as shown in \textbf{Example 1}, the model could refer to \tracespan{exceptions or error messages} in the trace (eg in \trace{2}) to \goodratspan{explain bugs in the code}.
Next, \textbf{Example 2} shows that the model could also leverage \tracespan{variable states} in the trace (\eg in \trace{2}) and compare them with the expected values to \goodratspan{locate the cause of bugs}.
Besides, the {\tt NO\_CHANGE} annotations for variables whose values are preserved after execution of a step could also help the model explain the execution process in the rationale (\eg {\tt (3)NO\_CHANGE} $\mapsto$ \textit{``the first sublist is already sorted''}).
Perhaps a more interesting scenario is when the model reasons over multiple steps of computation to track down the cause of a bug.
In \textbf{Example 3}, the model attempts to trace the computation of \tracespan{steps 2 - 4} in \trace{1} to explain why the sum is a float instead of an integer. 
Another example is \textbf{Example 4}, where the model summarizes the loop iterations in \tracespan{steps 2 - 9} of \trace{1} to explain the cause of the missing last element in the result list. 
Interestingly, while the model is able to reason over multiple steps of execution in its rationales, as the reasoning chain becomes longer, it is more likely to contain minor logical errors, as \badratspan{highlighted} in the rationales of Examples 3 and 4. 

Besides minor logical errors in CoT reasoning, we also identify several more pronounced error modes.
First, even if the NL rationale looks valid, the actual code fix might be incorrect (\textbf{Example 4}).
Second, for cases where there are multiple bugs in the program, the model might only focus on the more obvious bugs during reasoning (\eg \goodratspan{explaining the reason of {\tt IndexError}} in \trace{2} of \textbf{Example 5}), while ignoring the other more nuanced bugs that are buried in the trace (\eg \badratspan{Incorrect outputs} in \trace{1} and \trace{3} caused by zero-initialized {\tt freq\_arr}).
This could be especially problematic when there are hidden bugs \emph{not} captured by the trace, which could happen when a buggy program crashes in the middle of execution and the rest the of the code is not covered in the trace.
Since the model could be more reliant on trace information after fine-tuning, it might be more likely to miss such implicit bugs.
As an example, in \textbf{Example 6} the model only \goodratspan{noticed the {\tt TypeError} in the trace}, without spotting that the \badratspan{code that computes the end index in the output tuple} is off by one.

\clearpage
\newpage

\lstdefinestyle{examples}{
    backgroundcolor=\color{white},   
    commentstyle=\color{magenta},
    keywordstyle=\color{blue},
    numberstyle=\scriptsize\bfseries\color{codegray},
    stringstyle=\color{codepurple},
    basicstyle=\fontfamily{cmtt}\scriptsize,
    breakatwhitespace=false,         
    breaklines=true,    
    breakindent=0pt,
    captionpos=b,                    
    keepspaces=true,                 
    numbersep=5pt,                  
    showspaces=false,                
    showstringspaces=false,
    showtabs=false,                  
    tabsize=2,
    escapeinside={*@}{@*},
    belowskip=-\baselineskip,
    aboveskip=- 0.5\baselineskip,
}

\lstset{style=examples}

{\small
\begin{longtable}{p{\textwidth}}
\toprule
\textbf{Example 1:} An example where our model explains exceptions in the trace \\ \midrule
\textbf{Task:} Write a function to remove leading zeroes from an ip address \\
\textbf{Buggy Code:} \\
\begin{lstlisting}[language=Python,]
def removezero_ip(ip):
    ip_list = ip.split(".")
    for i in range(4):
        ip_list[i] = str(int(ip_list[i]))
    return ".".join(ip_list)
\end{lstlisting}\\
\textbf{Test Cases:} \\
\begin{lstlisting}[language=Python,]
assert removezero_ip("216.08.094.196")==('216.8.94.196') 
assert removezero_ip("12.01.024")==('12.1.24') 
assert removezero_ip("216.08.094.0196")==('216.8.94.196') 
\end{lstlisting} \\
\textbf{Traces:} \\
\begin{lstlisting}[language=Python]
# Trace 1:
def removezero_ip(ip):  # (0) ip='216.08.094.196';
    ip_list = ip.split(".")  # (1) ip_list=['216', '08', '094', '196'];
    for i in range(4):  # (2) i=0; (4) i=1; (6) i=2; ...; (10) NO_CHANGE;
        ip_list[i] = str(int(ip_list[i]))  # (3) NO_CHANGE; (5) ip_list=['216', '8', '094', '196']; (7) ip_list=['216', '8', '94', '196']; (9) NO_CHANGE;
    return ".".join(ip_list)  # (11) __return__='216.8.94.196';

assert removezero_ip("216.08.094.196")==('216.8.94.196')

# Trace 2:
def removezero_ip(ip):  # (0) ip='12.01.024';
    ip_list = ip.split(".")  # (1) ip_list=['12', '01', '024'];
    for i in range(4):  # (2) i=0; (4) i=1; (6) i=2; (8) i=3;
        ip_list[i] = str(int(ip_list[i]))  # (3) NO_CHANGE; (5) ip_list=['12', '1', '024']; (7) ip_list=['12', '1', '24']; (9)  *@\tracespan{\_\_exception\_\_ = IndexError('list index out of range')}@*
    return ".".join(ip_list)

assert removezero_ip("12.01.024")==('12.1.24')

# Trace 3:
def removezero_ip(ip):  # (0) ip='216.08.094.0196';
    ip_list = ip.split(".")  # (1) ip_list=['216', '08', '094', '0196'];
    for i in range(4):  # (2) i=0; (4) i=1; (6) i=2; ...; (10) NO_CHANGE;
        ip_list[i] = str(int(ip_list[i]))  # (3) NO_CHANGE; (5) ip_list=['216', '8', '094', '0196']; (7) ip_list=['216', '8', '94', '0196']; (9) ip_list=['216', '8', '94', '196'];
    return ".".join(ip_list)  # (11) __return__='216.8.94.196';

assert removezero_ip("216.08.094.0196")==('216.8.94.196')
\end{lstlisting}\\ \midrule
\textbf{Rationale and Code Solution:} (code fix is correct \ccmark) \\
\begin{lstlisting}
The problem spec asks to write a function to remove leading zeroes from an IP address. 

Now let's look at the program. The program first splits the IP address into a list of strings, and then iterates through the list to convert each string to an integer, and then back to a string. This will remove leading zeroes. *@\goodratspan{However, the problem is that it iterates through the list for 4 times, but the length of the}@*  *@\goodratspan{ list might not be 4. For example, when `ip="12.01.024"`, the length of the list is only 3. Thus when `i=3`,}@* *@\goodratspan{ it will throw an `IndexError`}@*.

Here is a simple fix:
\end{lstlisting}\\
\begin{lstlisting}[language=Python]
def removezero_ip(ip):
    ip_list = ip.split(".")
    for i in range(len(ip_list)):
        ip_list[i] = str(int(ip_list[i])).lstrip('0')
    return ".".join(ip_list)
\end{lstlisting} \\ \bottomrule
\end{longtable}
}

{\small
\begin{longtable}{p{\textwidth}}
\toprule
\textbf{Example 2:} An example where our model uses variable states to solve the task \\ \midrule
\textbf{Task:} Write a function to sort each sublist of strings in a given list of lists \\
\textbf{Buggy Code:} \\
\begin{lstlisting}[language=Python,]
def sort_sublists(words):
    words.sort(key = len)
    for w in words:
        w.sort()
    return words
\end{lstlisting}\\
\textbf{Test Cases:} \\
\begin{lstlisting}[language=Python,]
assert sort_sublists([['green', 'orange'], ['black', 'white'], ['white', 'black', 'orange']])==[['green', 'orange'], ['black', 'white'], ['black', 'orange', 'white']]
assert sort_sublists([['green', 'orange'], ['black'], ['green', 'orange'], ['white']])==[['green', 'orange'], ['black'], ['green', 'orange'], ['white']]
assert sort_sublists([['a','b'],['d','c'],['g','h'] , ['f','e']])==[['a', 'b'], ['c', 'd'], ['g', 'h'], ['e', 'f']]
\end{lstlisting} \\
\textbf{Traces:} \\
\begin{lstlisting}[language=Python]
# Trace 1:
def sort_sublists(words):  # (0) words=[['green', 'orange'], ['black', 'white'], ['white', 'black', 'orange']];
    words.sort(key = len)  # (1) NO_CHANGE;
    for w in words:  # (2) w=['green', 'orange']; (4) w=['black', 'white']; (6) w=['white', 'black', 'orange']; (8) NO_CHANGE;
        w.sort()  # (3) NO_CHANGE; (5) NO_CHANGE; (7) w=['black', 'orange', 'white'], words=[['green', 'orange'], ['black', 'white'], ['black', 'orange', 'white']];
    return words  # (9) __return__=[['green', 'orange'], ['black', 'white'], ['black', 'orange', 'white']];

assert sort_sublists([['green', 'orange'], ['black', 'white'], ['white', 'black', 'orange']])==[['green', 'orange'], ['black', 'white'], ['black', 'orange', 'white']]

# Trace 2:
def sort_sublists(words):  # (0) words=[['green', 'orange'], ['black'], ['green', 'orange'], ['white']];
    words.sort(key = len)  *@\tracespan{\# (1) words=[['black'], ['white'], ['green', 'orange'], ['green', 'orange']];}@*
    for w in words:  # (2) w=['black']; (4) w=['white']; (6) w=['green', 'orange']; ...; (10) NO_CHANGE;
        w.sort()  # (3) NO_CHANGE; (5) NO_CHANGE; (7) NO_CHANGE; (9) NO_CHANGE;
    return words  # (11) __return__=[['black'], ['white'], ['green', 'orange'], ['green', 'orange']];

assert sort_sublists([['green', 'orange'], ['black'], ['green', 'orange'], ['white']])==[['green', 'orange'], ['black'], ['green', 'orange'], ['white']]  # (12)  __exception__ = AssertionError()

# Trace 3:
def sort_sublists(words):  # (0) words=[['a', 'b'], ['d', 'c'], ['g', 'h'], ['f', 'e']];
    words.sort(key = len)  # (1) NO_CHANGE;
    for w in words:  # (2) w=['a', 'b']; (4) w=['d', 'c']; (6) w=['g', 'h']; ...; (10) NO_CHANGE;
        w.sort()  # (3) NO_CHANGE; (5) w=['c', 'd'], words=[['a', 'b'], ['c', 'd'], ['g', 'h'], ['f', 'e']]; (7) NO_CHANGE; (9) w=['e', 'f'], words=[['a', 'b'], ['c', 'd'], ['g', 'h'], ['e', 'f']];
    return words  # (11) __return__=[['a', 'b'], ['c', 'd'], ['g', 'h'], ['e', 'f']];

assert sort_sublists([['a','b'],['d','c'],['g','h'] , ['f','e']])==[['a', 'b'], ['c', 'd'], ['g', 'h'], ['e', 'f']]
\end{lstlisting}\\ \midrule
\textbf{Rationale and Code Solution:} (code fix is correct \ccmark)\\
\begin{lstlisting}
The problem spec asks to sort each sublist of strings in a given list of lists.

Now let's look at the program. The program first sorts the list of lists by the length of each sublist, and then sorts each sublist. The first assertion passes as the first sublist is already sorted, and the second sublist is also sorted, and the third sublist is sorted in the for loop. *@\goodratspan{The second assertion fails as the list of lists is sorted by the length of each sublist, and thus the}@* *@\goodratspan{return value is not the same as the input.}@* The third assertion passes as the first and the last sublists are already sorted, and the second and the third sublists are sorted in the for loop.

*@\goodratspan{The problem is that the program sorts the list of lists by the length of each sublist, which is not required}@* *@\goodratspan{ by the problem spec.}@* Thus we can simply remove the first line of the function.

Here is the corrected code:
\end{lstlisting}\\
\begin{lstlisting}[language=Python]
def sort_sublists(words):
    for w in words:
        w.sort()
    return words
\end{lstlisting} \\
\bottomrule
\end{longtable}
}

{\small
\begin{longtable}{p{\textwidth}}
\toprule
\textbf{Example 3:} An example where \ourpalm reasons over multiple steps of execution \\ \midrule
\textbf{Task:} Write a function to get the sum of a non-negative integer \\
\textbf{Buggy Code:} \\
\begin{lstlisting}[language=Python,]
def sum_digits(number):
    sum = 0
    while(number):
        sum += number %
        number = number / 10
    return sum
\end{lstlisting}\\
\textbf{Test Cases:} \\
\begin{lstlisting}[language=Python,]
assert sum_digits(345)==12
assert sum_digits(12)==3
assert sum_digits(97)==16
\end{lstlisting} \\
\textbf{Traces:} \\
\begin{lstlisting}[language=Python]
# Trace 1:
def sum_digits(number):  # (0) number=345;
    sum = 0  # (1) sum=0;
    while(number):
        sum += number %
        number = number / 10  # *@\tracespan{(3) number=34.5;}@* (5) number=3.45; (7) number=0.34500000000000003; ...; (655) number=0.0;
    return sum  # (656) __return__=13.33333333333333;

assert sum_digits(345)==12  # (657)  __exception__ = AssertionError()

# Trace 2:
def sum_digits(number):  # (0) number=12;
    sum = 0  # (1) sum=0;
    while(number):
        sum += number %
        number = number / 10  # (3) number=1.2; (5) number=0.12; (7) number=0.012; ...; (651) number=0.0;
    return sum  # (652) __return__=3.3333333333333335;

assert sum_digits(12)==3  # (653)  __exception__ = AssertionError()

# Trace 3:
def sum_digits(number):  # (0) number=97;
    sum = 0  # (1) sum=0;
    while(number):
        sum += number %
        number = number / 10  # (3) number=9.7; (5) number=0.97; (7) number=0.097; ...; (653) number=0.0;
    return sum  # (654) __return__=17.77777777777777;

assert sum_digits(97)==16  # (655)  __exception__ = AssertionError()
\end{lstlisting}\\ \midrule
\textbf{Rationale and Code Solution:} (code fix is correct \ccmark)\\
\begin{lstlisting}
The problem spec asks to write a function to get the sum of all digits in a non-negative integer.

Now let's look at the program. The program seems to be correct at the first glance. It first initializes the sum to 0, and then it uses a while loop to get each digit in the number and add it to the sum. However, if we look at the output of the program, we will find that the sum is not an integer. *@\goodratspan{This is because the program does not convert the number to an integer before it gets the digits. For example,}@* *@\goodratspan{ for the first assertion, when `number=345`, the first digit `5` is added to the sum and `sum=5`. Then `number`}@* *@\goodratspan{is updated to `34.5`,}@* *@\badratspan{and the second digit `4` is added to the sum}@* *@\goodratspan{and `sum=9.5`. And so on}@*.

Here is a fixed version:
\end{lstlisting}\\
\begin{lstlisting}[language=Python]
def sum_digits(number):
    number = int(number)
    sum = 0
    while(number):
        sum += number %
        number = number // 10
    return sum
\end{lstlisting} \\
\bottomrule
\end{longtable}
}

{\small
\begin{longtable}{p{\textwidth}}
\toprule
\textbf{Example 4:} An example where \ourpalm reasons over multiple steps of execution \\ \midrule
\textbf{Task:} Write a function to convert tuple into list by adding the given string after every element. \\
\textbf{Buggy Code:} \\
\begin{lstlisting}[language=Python]
def add_str(test_tuple, add_str):
    res = list(test_tuple)
    for i in range(1,len(test_tuple)):
        res.insert(i*2-1,add_str)
    return res
\end{lstlisting}\\
\textbf{Test Cases:} \\
\begin{lstlisting}[language=Python,]
assert add_str((5, 6, 7, 4, 9) , "FDF") == [5, 'FDF', 6, 'FDF', 7, 'FDF', 4, 'FDF', 9, 'FDF']
assert add_str((7, 8, 9, 10) , "PF") == [7, 'PF', 8, 'PF', 9, 'PF', 10, 'PF']
assert add_str((11, 14, 12, 1, 4) , "JH") == [11, 'JH', 14, 'JH', 12, 'JH', 1, 'JH', 4, 'JH']
\end{lstlisting} \\
\textbf{Traces:} \\
\begin{lstlisting}[language=Python]
# Trace 1:
def add_str(test_tuple, add_str):
    # Write your code here
    res = list(test_tuple)  # (1) res=[5, 6, 7, 4, 9];
    for i in range(1,len(test_tuple)):  # *@\tracespan{(2) i=1; (4) i=2; (6) i=3; ...;}@* (10) NO_CHANGE;
        res.insert(i*2-1,add_str)  # *@\tracespan{(3) res=[5, 'FDF', 6, 7, 4, 9]; (5) res=[5, 'FDF', 6, 'FDF', 7, 4, 9]; }@*
        *@\tracespan{(7) res=[5, 'FDF', 6, 'FDF', 7, 'FDF', 4, 9]; (9) res=[5, 'FDF', 6, 'FDF', 7, 'FDF', 4, 'FDF', 9];}@*
    return res  # (11) __return__=[5, 'FDF', 6, 'FDF', 7, 'FDF', 4, 'FDF', 9];

assert add_str((5, 6, 7, 4, 9) , "FDF") == [5, 'FDF', 6, 'FDF', 7, 'FDF', 4, 'FDF', 9, 'FDF']  # (12)  __exception__ = AssertionError()

# Trace 2:
def add_str(test_tuple, add_str):
    # Write your code here
    res = list(test_tuple)  # (1) res=[7, 8, 9, 10];
    for i in range(1,len(test_tuple)):  # (2) i=1; (4) i=2; (6) i=3; (8) NO_CHANGE;
        res.insert(i*2-1,add_str)  # (3) res=[7, 'PF', 8, 9, 10]; (5) res=[7, 'PF', 8, 'PF', 9, 10]; (7) res=[7, 'PF', 8, 'PF', 9, 'PF', 10];
    return res  # (9) __return__=[7, 'PF', 8, 'PF', 9, 'PF', 10];

assert add_str((7, 8, 9, 10) , "PF") == [7, 'PF', 8, 'PF', 9, 'PF', 10, 'PF']  # (10)  __exception__ = AssertionError()

# Trace 3:
def add_str(test_tuple, add_str):
    # Write your code here
    res = list(test_tuple)  # (1) res=[11, 14, 12, 1, 4];
    for i in range(1,len(test_tuple)):  # (2) i=1; (4) i=2; (6) i=3; ...; (10) NO_CHANGE;
        res.insert(i*2-1,add_str)  # (3) res=[11, 'JH', 14, 12, 1, 4]; (5) res=[11, 'JH', 14, 'JH', 12, 1, 4]; (7) res=[11, 'JH', 14, 'JH', 12, 'JH', 1, 4]; (9) res=[11, 'JH', 14, 'JH', 12, 'JH', 1, 'JH', 4];
    return res  # (11) __return__=[11, 'JH', 14, 'JH', 12, 'JH', 1, 'JH', 4];

assert add_str((11, 14, 12, 1, 4) , "JH") == [11, 'JH', 14, 'JH', 12, 'JH', 1, 'JH', 4, 'JH']  # (12)  __exception__ = AssertionError()
\end{lstlisting}\\ \midrule
\textbf{Rationale and Code Solution:} (code fix is wrong \xxmark)\\
\begin{lstlisting}
The problem spec asks to convert a tuple to a list and add a given string after every element.

Now let's look at the program. The program first converts the tuple to a list, which is good. Then it iterates from 1 to the length of the list, and inserts the given string at index `i*2-1`. This seems correct at the first glance, but *@\goodratspan{the problem is that it will not add the string after the last element. For example, when}@* *@\goodratspan{ `test\_tuple=(5, 6, 7, 4, 9)`, the length of the list is 5, and the for loop will iterate from 1 to 4. Thus the}@* *@\goodratspan{ string will be inserted at index 1, 3, 5, and 7,}@* *@\badratspan{which corresponds to after the second, fourth, and sixth}@* *@\badratspan{ element.}@* *@\goodratspan{The string will not be inserted after the last element (i.e., the ninth element).}@*

Here is a simple fix:
\end{lstlisting}\\
\begin{lstlisting}[language=Python]
def add_str(test_tuple, add_str):
    res = list(test_tuple)
    for i in range(len(test_tuple)):
        res.insert(i*2,add_str)
    return res
\end{lstlisting} \\
\bottomrule
\end{longtable}
}

{\small
\begin{longtable}{p{\textwidth}}
\toprule
\textbf{Example 5:} An example where \ourpalm failed to identify all the bugs in the program \\ \midrule
\textbf{Task:} Write a python function to find the difference between highest and least frequencies in a given array. \\
\textbf{Buggy Code:} \\
\begin{lstlisting}[language=Python]
def find_Diff(nums, n):
    freq_arr = [0 for i in range(n)]
    for i in range(len(nums)):
        freq_arr[nums[i]] = freq_arr[nums[i]] + 1
    return max(freq_arr) - min(freq_arr)
\end{lstlisting}\\
\textbf{Test Cases:} \\
\begin{lstlisting}[language=Python,]
assert find_Diff([1,1,2,2,7,8,4,5,1,4],10) == 2
assert find_Diff([1,7,9,2,3,3,1,3,3],9) == 3
assert find_Diff([1,2,1,2],4) == 0
\end{lstlisting} \\
\textbf{Traces:} \\
\begin{lstlisting}[language=Python]
# Trace 1:
def find_Diff(nums, n):  # (0) n=10, nums=[1, 1, 2, 2, 7, 8, 4, 5, 1, 4]; (1) .0=<range_iterator object>;
    freq_arr = [0 for i in range(n)]  # (2) i=0; (3) i=1; (4) i=2; ...; (13) freq_arr=[0, 0, 0, 0, 0, 0, 0, 0, 0, 0];
    for i in range(len(nums)):  # (14) i=0; (16) i=1; (18) i=2; ...; (34) NO_CHANGE;
        freq_arr[nums[i]] = freq_arr[nums[i]] + 1  # (15) freq_arr=[0, 1, 0, 0, 0, 0, 0, 0, 0, 0]; (17) freq_arr=[0, 2, 0, 0, 0, 0, 0, 0, 0, 0]; (19) freq_arr=[0, 2, 1, 0, 0, 0, 0, 0, 0, 0]; ...; (33) freq_arr=[0, 3, 2, 0, 2, 1, 0, 1, 1, 0];
    return max(freq_arr) - min(freq_arr)  # (35) __return__=3;

assert find_Diff([1,1,2,2,7,8,4,5,1,4],10) == 2  # *@\badratspan{(36)  \_\_exception\_\_ = AssertionError()}@*

# Trace 2:
def find_Diff(nums, n):  # (0) n=9, nums=[1, 7, 9, 2, 3, 3, 1, 3, 3]; (1) .0=<range_iterator object>;
    freq_arr = [0 for i in range(n)]  # (2) i=0; (3) i=1; (4) i=2; ...; (12) freq_arr=[0, 0, 0, 0, 0, 0, 0, 0, 0];
    for i in range(len(nums)):  # (13) i=0; (15) i=1; (17) i=2;
        freq_arr[nums[i]] = freq_arr[nums[i]] + 1  # (14) freq_arr=[0, 1, 0, 0, 0, 0, 0, 0, 0]; (16) freq_arr=[0, 1, 0, 0, 0, 0, 0, 1, 0]; (18)  __exception__ = IndexError('list index out of range')
    return max(freq_arr) - min(freq_arr)

assert find_Diff([1,7,9,2,3,3,1,3,3],9) == 3

# Trace 3:
def find_Diff(nums, n):  # (0) n=4, nums=[1, 2, 1, 2]; (1) .0=<range_iterator object>;
    freq_arr = [0 for i in range(n)]  # (2) i=0; (3) i=1; (4) i=2; ...; (7) freq_arr=[0, 0, 0, 0];
    for i in range(len(nums)):  # (8) i=0; (10) i=1; (12) i=2; ...; (16) NO_CHANGE;
        freq_arr[nums[i]] = freq_arr[nums[i]] + 1  # (9) freq_arr=[0, 1, 0, 0]; (11) freq_arr=[0, 1, 1, 0]; (13) freq_arr=[0, 2, 1, 0]; (15) freq_arr=[0, 2, 2, 0];
    return max(freq_arr) - min(freq_arr)  # (17) __return__=2;

assert find_Diff([1,2,1,2],4) == 0  # *@\badratspan{(18) \_\_exception\_\_ = AssertionError()}@*
\end{lstlisting}\\ \midrule
\textbf{Rationale and Code Solution:} (code fix is wrong \xxmark)\\
\begin{lstlisting}
The problem spec asks to find the difference between the highest and the least frequencies in a given array.

Now let's look at the program. The program first creates a frequency array of length `n`, and then iterates through the input array to count the frequency of each number. It then returns the difference between the maximum and the minimum frequencies in the frequency array.

The program looks good at the first glance. *@\goodratspan{However, the problem is that the length of the frequency array is `n`, which is the length of the input array.}@* *@\goodratspan{ However, the numbers in the input array can be larger than `n`, and thus the frequency array will not be able}@* *@\goodratspan{ to accommodate all the numbers in the input array.}@* For example, in the second assertion, the input array is `[1, 7, 9, 2, 3, 3, 1, 3, 3]`, and the length of the frequency array is also `9`. When `nums[i]=9`, it will try to access `freq_arr[9]`, which is out of bound, and thus an `IndexError` is thrown.

Here is a simple fix:
\end{lstlisting}\\
\begin{lstlisting}[language=Python]
def find_Diff(nums, n):
    freq_arr = {}
    for i in range(len(nums)):
        if nums[i] in freq_arr:
            freq_arr[nums[i]] += 1
        else:
            freq_arr[nums[i]] = 1
    max_freq = 0
    min_freq = 0
    for num, freq in freq_arr.items():
        if freq > max_freq:
            max_freq = freq
        if freq < min_freq or min_freq == 0:
            min_freq = freq
    return max_freq - min_freq
\end{lstlisting} \\
\bottomrule
\end{longtable}
}

{\small
\begin{longtable}{p{\textwidth}}
\toprule
\textbf{Example 6:} An example where \ourpalm failed to identify a bug not revealed in the trace \\ \midrule
\textbf{Task:} Write a function to search a literals string in a string and also find the location within the original string where the pattern occurs. \\
\textbf{Buggy Code:} \\
\begin{lstlisting}[language=Python]
def search_literal(pattern, text):
    match = tuple(text.find(pattern), *@\badratspan{text.find(pattern) + len(pattern) - 1}@*)

    return match
\end{lstlisting}\\
\textbf{Test Cases:} \\
\begin{lstlisting}[language=Python,]
assert search_literal('python','python programming language')==(0,6)
assert search_literal('programming','python programming language')==(7,18)
assert search_literal('language','python programming language')==(19,27)
\end{lstlisting} \\
\textbf{Traces:} \\
\begin{lstlisting}[language=Python]
# Trace 1:
def search_literal(pattern, text):
    match = tuple(text.find(pattern), text.find(pattern) + len(pattern) - 1)  # (1)  __exception__ = TypeError('tuple expected at most 1 argument, got 2')
    
    return match

assert search_literal('python','python programming language')==(0,6)

# Trace 2:
def search_literal(pattern, text):
    match = tuple(text.find(pattern), text.find(pattern) + len(pattern) - 1)  # (1)  __exception__ = TypeError('tuple expected at most 1 argument, got 2')
    
    return match

assert search_literal('programming','python programming language')==(7,18)

# Trace 3:
def search_literal(pattern, text):
    match = tuple(text.find(pattern), text.find(pattern) + len(pattern) - 1)  # (1)  __exception__ = TypeError('tuple expected at most 1 argument, got 2')
    
    return match

assert search_literal('language','python programming language')==(19,27)
\end{lstlisting}\\ \midrule
\textbf{Rationale and Code Solution:} (code fix is wrong \xxmark)\\
\begin{lstlisting}
Now let's look at the program. The program looks good at the first glance as `text.find(pattern)` is the correct way to find the location of a literal string in a given string. But the problem is that it does not use the `tuple()` function correctly. The `tuple()` function takes a single iterable as input, and returns a tuple with the elements from the iterable. *@\goodratspan{But in the program, it takes two arguments, which will throw a}@* *@\goodratspan{ `TypeError` of `tuple expected at most 1 argument, got 2`.}@* 

Here is a simple fix:
\end{lstlisting}\\
\begin{lstlisting}[language=Python]
def search_literal(pattern, text):
    match = tuple(text.find(pattern),)
    match = match + (text.find(pattern) + len(pattern) - 1,)
    
    return match
\end{lstlisting} \\
\bottomrule
\end{longtable}
}

\newpage
\section{Full Prompt}
\label{sec:full-prompt}
Here we show the instruction and all three few-shot examples used in this work. The rationales and fixes in the few-shot examples are hand annotated by the authors. Each exemplar has three unit tests and their traces.

\usemintedstyle{xcode}
\makeatletter
\AtBeginEnvironment{minted}{\dontdofcolorbox}
\def\dontdofcolorbox{\renewcommand\fcolorbox[4][]{##4}}
\makeatother

\setminted{xleftmargin=2em}

\begin{promptfence}
\label{lst:prompt:sbs:e1}
\begin{minted}[xleftmargin=2em,linenos, breaklines, firstnumber=last,escapeinside=||]{python}

|We are doing pair programming, I will give you the problem definition and my implementation, which does not pass all test cases. And you need to help me to fix my code.|

# Example 1

Here is the specification in natural language and the program that I wrote:
```
# Write a function to find the similar elements from the given two tuple lists.
def similar_elements(t1, t2): 
	s1 = set(t1) 
	s2 = set(t2) 
	return s1.intersection(s2)

```
Here are the assertions I need to pass:
```
assert similar_elements((3, 4, 5, 6),(5, 7, 4, 10)) == (4, 5)
assert similar_elements((1, 2, 3, 4),(5, 4, 3, 7)) == (3, 4)
assert similar_elements((11, 12, 14, 13),(17, 15, 14, 13)) == (13, 14)
```

|Here is a *cheatsheet* for the gold execution trace for each assertions, use the information here but pretend you did all the reasoning about execution yourself.|

#################### cheatsheet starts ####################
def similar_elements(t1, t2):   # (0) t1=(3, 4, 5, 6), t2=(5, 7, 4, 10); 
	s1 = set(t1)   # (1) s1={3, 4, 5, 6}; 
	s2 = set(t2)   # (2) s2={10, 4, 5, 7}; 
	return s1.intersection(s2)  # (3) __return__={4, 5}


assert similar_elements((3, 4, 5, 6),(5, 7, 4, 10)) == (4, 5)  #
--------------------
def similar_elements(t1, t2):   # (0) t1=(1, 2, 3, 4), t2=(5, 4, 3, 7); 
	s1 = set(t1)   # (1) s1={1, 2, 3, 4}; 
	s2 = set(t2)   # (2) s2={3, 4, 5, 7}; 
	return s1.intersection(s2)  # (3) __return__={3, 4}


assert similar_elements((1, 2, 3, 4),(5, 4, 3, 7)) == (3, 4)  #
--------------------
def similar_elements(t1, t2):   # (0) t1=(11, 12, 14, 13), t2=(17, 15, 14, 13); 
	s1 = set(t1)   # (1) s1={11, 12, 13, 14}; 
	s2 = set(t2)   # (2) s2={17, 13, 14, 15}; 
	return s1.intersection(s2)  # (3) __return__={13, 14}


assert similar_elements((11, 12, 14, 13),(17, 15, 14, 13)) == (13, 14)  #
#################### cheatsheet ends ####################

|Given the above information, can you tell me why my code is wrong, and also provide a fixed version?|

|The problem spec asks to find the similar elements in the two tuples, from the assertion we can see that it essentially wants the common elements from the two tuples. It also seems that the common elements should be returned in a tuple with ascending order.|

Now let's look at the program. While it is correct and efficient to use `intersection()` to find the common elements between two sets, it will also return a set. Take the first assertion as an example, `s1` will first be converted to a set like `{3, 4, 5, 6}` and similarly for `s2`, then the return value will be `{4, 5}`. Though the elements are right, the expected return value is a tuple and not a set. Thus we only need to further convert the sets to tuples using `tuple()`.

Here is the corrected code:
```
def similar_elements(t1, t2): 
	s1 = set(t1) 
	s2 = set(t2) 
	return tuple(s1.intersection(s2))
```

# Example 2

Here is the specification in natural language and the program that I wrote:
```
# Write a python function to identify non-prime numbers.
def is_not_prime(n):
  if n==1:
    return False
  else:
    for i in range(2,n//2+1):
      if n%
        return False
      else:
        return True
```
Here are the assertions I need to pass:
```
assert is_not_prime(2) == False
assert is_not_prime(10) == True
assert is_not_prime(35) == True
```

|Here is a *cheatsheet* for the gold execution trace for each assertions, use the information here but pretend you did all the reasoning about execution yourself.|

#################### cheatsheet starts ####################
def is_not_prime(n):  # (0) n=2; 
  if n==1:  # 
    return False
  else:
    for i in range(2,n//2+1):  # (1) __return__=None
      if n%
        return False
      else:
        return True


assert is_not_prime(2) == False  #
--------------------
def is_not_prime(n):  # (0) n=10; 
  if n==1:  # 
    return False
  else:
    for i in range(2,n//2+1):  # (1) i=2; 
      if n%
        return False  # (2) __return__=False
      else:
        return True


assert is_not_prime(10) == True  #
--------------------
def is_not_prime(n):  # (0) n=35; 
  if n==1:  # 
    return False
  else:
    for i in range(2,n//2+1):  # (1) i=2; 
      if n%
        return False
      else:
        return True  # (2) __return__=True


assert is_not_prime(35) == True  #
#################### cheatsheet ends ####################

|Given the above information, can you tell me why my code is wrong, and also provide a fixed version?|

|The problem spec asks to write a Python function that identifies non-prime numbers. Non-prime numbers are numbers that are divisible by numbers other than 1 and itself. Note that 0 and 1 are also non-prime numbers.|

Now let's check the program. Let's run the program through some examples. For the first assertion, when `n=2`, it will enter the `else` branch, and then in the for loop, since the condition is `i in range(2, n//2+1)`, which is `range(2, 2)`, it will actually not enter the loop since the condition is not met in the first place. And then it will reach the end of the function, and the return value will be `None`. And for the third assertion, when `n=35`, it also enters the `else` branch, and it then enters the loop with `i=2` as `n//2+1=18 > 2`. Then in the next step, since `n%

While it is not obvious how to correct this program, another way to implement the `is_not_prime()` function is just to first create a helper function to identify prime numbers as `is_prime()`, and then simply `return not is_prime(x)`.

Here is a re-implementation:
```
def is_prime(x):
  if x <= 1:
    return False
  else:
    for i in range(2,x):
      if x%
        return False
    return True

def is_not_prime(x):
  return not is_prime(x)
```

# Example 3

Here is the specification in natural language and the program that I wrote:
```
# Write a function to find the largest integers from a given list of numbers using heap queue algorithm.
def heap_queue_largest(list1,n):
  largest=heapq.nlargest(n,list1)
  return largest

```
Here are the assertions I need to pass:
```
assert heap_queue_largest( [25, 35, 22, 85, 14, 65, 75, 22, 58],3)==[85, 75, 65] 
assert heap_queue_largest( [25, 35, 22, 85, 14, 65, 75, 22, 58],2)==[85, 75] 
assert heap_queue_largest( [25, 35, 22, 85, 14, 65, 75, 22, 58],5)==[85, 75, 65, 58, 35]
```

|Here is a *cheatsheet* for the gold execution trace for each assertions, use the information here but pretend you did all the reasoning about execution yourself.|

#################### cheatsheet starts ####################
def heap_queue_largest(list1,n):  # (0) list1=[25, 35, 22, 85, 14, 65, 75, 22, 58], n=3; 
  largest=heapq.nlargest(n,list1)  # (1)  __exception__ = NameError("name 'heapq' is not defined")
  return largest


assert heap_queue_largest( [25, 35, 22, 85, 14, 65, 75, 22, 58],3)==[85, 75, 65]
--------------------
def heap_queue_largest(list1,n):  # (0) list1=[25, 35, 22, 85, 14, 65, 75, 22, 58], n=2; 
  largest=heapq.nlargest(n,list1)  # (1)  __exception__ = NameError("name 'heapq' is not defined")
  return largest


assert heap_queue_largest( [25, 35, 22, 85, 14, 65, 75, 22, 58],2)==[85, 75]
--------------------
def heap_queue_largest(list1,n):  # (0) list1=[25, 35, 22, 85, 14, 65, 75, 22, 58], n=5; 
  largest=heapq.nlargest(n,list1)  # (1)  __exception__ = NameError("name 'heapq' is not defined")
  return largest


assert heap_queue_largest( [25, 35, 22, 85, 14, 65, 75, 22, 58],5)==[85, 75, 65, 58, 35]
#################### cheatsheet ends ####################

|Given the above information, can you tell me why my code is wrong, and also provide a fixed version?|

The problem spec asks to find `n` largest integers in a given list. It also suggests that the heap queue shall be used.

Now let's look at the program. The program looks good at the first glance as `heapq.nlargest()` is the correct way to get the `n` largest integers from an iterable (e.g., a list). But the problem is that it does not import the `heapq` package first. Thus all three assertions will fail at the first line of the function, and throw a `NameError` of `name 'heapq' is not defined`. 

Here is a simple fix:
```
import heapq
def heap_queue_largest(list1,n):
  largest=heapq.nlargest(n,list1)
  return largest
```
\end{minted}
\end{promptfence}

\end{document}